\pdfoutput=1
\documentclass{article}
\usepackage{arxiv}
\usepackage[T1]{fontenc}
\usepackage[utf8]{inputenc}
\usepackage{textcomp,microtype}
\usepackage{amsmath,amssymb,booktabs,graphicx,tabularx,array,seqsplit}
\usepackage{xcolor,url,placeins,float}
\usepackage[authoryear,round]{natbib}
\usepackage{hyperref}

\hypersetup{colorlinks=true,citecolor=blue!55!black,linkcolor=blue!55!black,urlcolor=blue!55!black,
 pdftitle={Ranking Agreement Does Not Reveal Paragraph-Edit Agreement in Two Released Paper Acceptance Predictors},
 pdfauthor={Shaoliang Yang and Jun Wang}}
\newcommand{\Surr}{\ensuremath{S}}
\newcommand{\Tone}{\ensuremath{T_1}}
\newcommand{\Ttwo}{\ensuremath{T_2}}
\newcommand{\CI}[3]{\ensuremath{#1\;[#2,\,#3]}}

\renewcommand{\shorttitle}{Ranking Agreement Does Not Reveal Edit Agreement}
\renewcommand{\headeright}{A Preprint}
\title{Ranking Agreement Does Not Reveal Paragraph-Edit Agreement in Two Released Paper Acceptance Predictors}
\author{Shaoliang Yang \quad Jun Wang\thanks{Corresponding author: \texttt{jwang22@scu.edu}.}\\
\normalfont Department of Mechanical Engineering, Santa Clara University\\
\normalfont 500 El Camino Real, Santa Clara, CA 95053, USA}
\date{November 2026}
\newcommand{\posthoctablebarrier}{}
\begin{document}
\maketitle
\begin{abstract}
A fast acceptance classifier could select revisions for a slower evaluator. We ask whether agreement on existing papers supports that use. We compare a classifier of word and document features with two released text-only acceptance predictors fine-tuned from the same language model with three billion parameters. For 198 papers from the International Conference on Learning Representations, we compare model rankings of the original papers and of paragraph rewrites screened automatically for unchanged facts, claim strength, and attribution. The classifier ranks original papers closely with one predictor and moderately with the other, but shows no detectable agreement with either on which rewrites raise scores. The tests specified in advance used reduced numerical precision. Rounding error dominated the small score responses. We therefore rescored the released weights in full 32-bit precision as a post hoc check; edit agreement remained near zero. At this precision, when each predictor chooses among five rewrites per paper, its average raw score is under two points out of 100 higher than with random choice. The evidence covers presentation rewrites and two open predictors from one family; transfer to costly evaluators remains untested. Revision selectors should therefore be checked against the intended evaluator on the revisions they rank.
\end{abstract}
\section{Introduction}
\label{sec:introduction}
Predicting whether a paper will be accepted and deciding how to revise it are different tasks. A predictor may recognize features associated with acceptance across papers without identifying useful changes to any one paper. For example, two scorers could rank the same manuscript highly yet favor different rewrites of its opening paragraph. A revision tool needs evidence about this second comparison: which changes it rewards.

We study two text-only models from PaperScorer, which predicts acceptance from manuscript text and page images and presents its outputs as calibrated acceptance estimates \citep{paperscorer}. The release reports revisions guided by a model's own feedback. We instead select revisions with a separate, faster classifier and then measure how the two models score them.

The fast classifier, called the surrogate \Surr, uses words and document features. The two acceptance models, called the targets \Tone{} and \Ttwo, share a base language model but were trained on different paper collections with different prompts. Each target has three billion parameters. We select edits using \Surr{} alone; neither target provides feedback during generation or selection.

PaperScorer was the first released predictor we considered, rather than the winner of a comparison among candidates. Its public test split supplies the same original papers for all three scorers, while the two targets share a base model. Text-only inputs let paragraph edits map directly onto the scored input. We use papers from the International Conference on Learning Representations (ICLR); the page-image and larger released models are outside the tested pair.

The open targets let us model a setting in which an author cannot query a costly evaluator repeatedly and uses a cheaper scorer to select revisions. This is a hypothetical use, rather than a documented revision tool. The tested targets are themselves cheap to query; we score them to measure their responses, and transfer to costly evaluators remains untested.

Original-paper agreement may not be sufficient for this use. Two generative reviewers can agree on original-paper scores while agreeing much less on the gains they assign to the same whole-paper revisions \citep{gainagreement}. Presentation-based attacks also show that wording can be optimized and that some attacks transfer between reviewers \citep{repackaging,abstractattack}. More generally, models that perform alike on held-out data can generalize differently under distribution shift \citep{damour2022,mccoy2020}. Unlike whole-paper revisions, single-paragraph rewrites let us compare several alternatives within each paper. We ask whether a classifier that tracks a target's paper rankings also tracks its responses to those alternatives. We measure agreement about changes, without treating a score response as desirable.

The same distinction matters for large language model (LLM) judges \citep{mtbench,geval} and learned reward models used to optimize text \citep{ziegler,summarizehf}. Their agreement on fixed texts does not establish that they reward the same revisions. Our evidence motivates corresponding tests without establishing how these systems would perform.

\begin{figure}[p]
\centering
\includegraphics[width=0.92\linewidth]{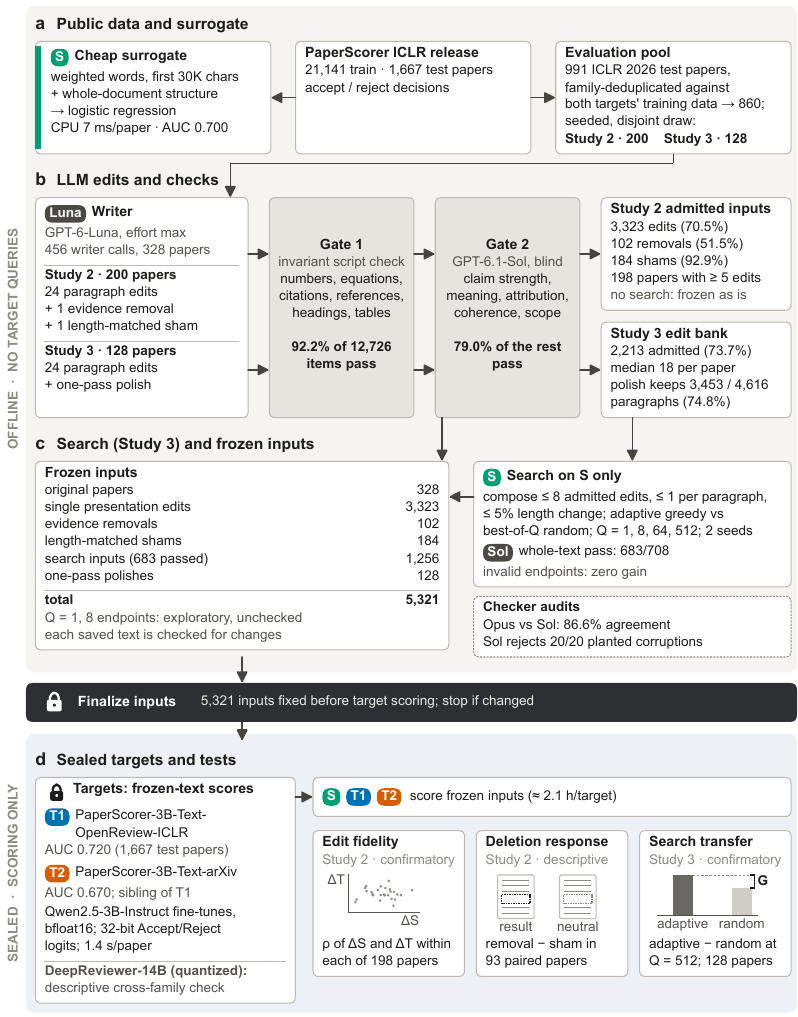}
\caption{Study design. \Surr{} is the fast surrogate; \Tone{} and \Ttwo{} are acceptance models. Paragraph edits pass invariant and blind semantic checks before surrogate-only search. All 5,321 inputs are fixed before target scoring; earlier original-paper scores are excluded from sampling and search. The human audit follows scoring. The miniature plots are schematic. ICLR denotes the International Conference on Learning Representations; AUC, area under the receiver operating characteristic curve; CPU, central processing unit. LLM denotes large language model; bfloat16 is a reduced-precision numerical format. In the schematic, $\Delta$ denotes a score change, $\rho$ Spearman's rank correlation \citep{spearman1904}, $G$ the adaptive-minus-random target gain, and $Q$ the maximum number of surrogate evaluations; Opus and Sol denote Claude Opus~5.5 and GPT-6.1-Sol.}
\label{fig:pipeline}
\end{figure}

Figure~\ref{fig:pipeline} maps the three studies. Study~1 compares acceptance prediction and score rankings on original ICLR papers. Study~2 compares rankings of paragraph rewrites that pass automatic preservation checks. Study~3 compares combinations chosen by adaptive and random search, both selecting with the surrogate's scores.

The classifier's agreement on original papers does not extend to agreement about which paragraph rewrites raise the predictors' scores. Because the score changes are small, the study bounds agreement for these checked rewrites; it does not establish whether any model's edit preferences are valid.

\section{Evaluators and study design}
\label{sec:methods}
\subsection{Which scorers are compared?}
The surrogate \Surr{} is logistic regression trained on PaperScorer ICLR acceptance labels: 21,141 training papers and 844 validation papers. It combines term frequency--inverse document frequency (TF--IDF) word features from the first 30,000 characters with whole-document counts of length, figures, tables, equations, headings, and citations, plus title and abstract length. The word weights favor terms common within a paper but rare across papers. The surrogate is fitted to these labels only: it is not distilled from target scores and is separate from the feature baseline reported with PaperScorer. Feature settings and regularization are selected by validation area under the receiver operating characteristic curve (AUC), without edit outcomes. Supplementary Section~\ref{app:surrogate} gives the full configuration.

The acceptance models are PaperScorer-3B-Text-OpenReview-ICLR (\Tone) \citep{cardiclr} and PaperScorer-3B-Text-arXiv (\Ttwo) \citep{cardarxiv}, both released with PaperScorer \citep{paperscorer}. Both released models are fixed fine-tunes of Qwen2.5-3B-Instruct \citep{qwen25}, using the same base checkpoint and training approach. \Tone{} is trained on ICLR papers; \Ttwo{} uses arXiv papers associated with several machine learning venues. 

\begin{samepage}
Each uses its released prompt and scoring rule. The raw score is the probability of the acceptance token divided by the sum of the acceptance and rejection probabilities (Supplementary Section~\ref{app:surrogate}). The release also maps this score to a calibrated acceptance estimate by temperature scaling with a term for the prior acceptance rate; we use the raw score, and because that mapping is monotone, within-paper rankings are the same under either.
\end{samepage}

Each fixed input receives one score from each target on an RTX 4090 graphics processing unit (GPU). Inputs longer than the release's training cutoff of 24,480 tokens are truncated. Table~\ref{tab:precision_main} separates the four scoring paths. Our tests use Transformers, a model-inference library \citep{transformers}; the release uses vLLM, a model-serving library \citep{vllm}. The release does not set a precision, so vLLM's current default uses float16 for these 32-bit checkpoints \citep{vllmdocs}. Supplementary Section~\ref{app:reliability} gives the precision comparisons.

\begin{table}[!b]
\centering
\caption{Original-paper prediction and scoring cost. Area under the receiver operating characteristic curve (AUC) is measured on the same 1,667 released International Conference on Learning Representations (ICLR) test papers, including 834 accepted papers. The split is held out for \Surr{} and \Tone{}, but 160 titles also occur in \Ttwo's training corpus; its row is contextual, rather than a fully held-out validation estimate. Brackets are nominal 95\% bootstrap confidence intervals (CIs); rank agreement uses Spearman's correlation \citep{spearman1904}. The table compares each scorer with \Surr; the correlation between \Tone{} and \Ttwo{} is 0.626. On its own arXiv test split, \Ttwo's AUC is 0.813. Latencies are medians for scoring already-extracted text, one input at a time, in this study's implementation rather than the release recipe: \Surr{} uses a central processing unit (CPU), while the targets use one RTX 4090 graphics processing unit (GPU). CPU timing uses the first 200 benchmark papers; target timing uses all 5,321 frozen inputs. This is not a common-workload hardware benchmark.}
\label{tab:context}
\begin{tabular}{lccc}
\toprule
Evaluator & AUC [95\% CI] & Spearman with \Surr & Time per paper \\
\midrule
\Surr & \CI{0.700}{0.675}{0.725} & 1.000 & 6.98 ms \\
\Tone & \CI{0.720}{0.695}{0.743} & 0.731 & 1.406 s \\
\Ttwo & \CI{0.670}{0.644}{0.695} & 0.547 & 1.405 s \\
\bottomrule
\end{tabular}
\end{table}

\begin{table}[tb]
\centering
\small
\caption{Target-scoring implementations. The released checkpoint files store 32-bit weights. Forward precision is the number format used in model calculations; our three scored implementations read the acceptance/rejection decision in 32-bit. The exact release recipe remains untested.}
\label{tab:precision_main}
\begin{tabularx}{\linewidth}{@{}p{0.24\linewidth}p{0.27\linewidth}X@{}}
\toprule
Purpose & Forward calculations & Analysis status \\
\midrule
Planned tests & bfloat16, Transformers & Four pre-specified tests \\
Released-weight check & 32-bit, Transformers & Post hoc; basis for interpreting the released weights \\
Release-recipe proxy & float16, Transformers & Post hoc approximation \\
Release recipe & float16, vLLM default & Not run \\
\bottomrule
\end{tabularx}
\end{table}

Table~\ref{tab:context} motivates \Surr{} as a proxy for \Tone: their original-paper rankings agree strongly, and their AUCs differ by 0.020. Supplementary Section~\ref{app:postreview} gives the paired AUC comparisons for both precisions.

\Ttwo{} supplies a second comparison with weaker original-paper agreement with \Surr{} (Table~\ref{tab:context}). It scores ICLR text with its released arXiv prompt, and some benchmark titles occur in its training corpus. On its own arXiv test split, our implementation reproduces the reported AUC (0.813 against 0.812 \citep{cardarxiv}; Supplementary Section~\ref{app:postreview}).

These qualifications make \Surr--\Ttwo{} a weaker premise for revision selection. We therefore treat \Surr--\Tone{} as the main comparison in our interpretation. This is a post hoc designation; the pre-specified family weighs both targets alike.

\subsection{Which papers and inputs are scored?}
Study 1 uses all 1,667 released ICLR test papers. From the 2026 papers, a seeded selection draws 200 for single edits (Study 2) and a disjoint 128 for search (Study 3). Before selection, we exclude pilot papers and likely training-paper relatives identified by title matches and title-plus-abstract similarity. Table~\ref{tab:cohort_main} maps the selected and analyzed sets. Supplementary Section~\ref{app:surrogate} gives the exclusions, Table~\ref{tab:cohorts} maps the sets, and Section~\ref{app:protocol} gives the seeds. These exclusions filter identified training families, not unknown base-model pretraining exposure.

\begin{table}[tb]
\centering
\small
\caption{How the analysis sets relate. Study 2 and Study 3 use disjoint papers. The edit rows are nested, so their counts must not be added. The edit-ranking analysis requires at least five admitted edits and nonconstant scores for both compared evaluators. Supplementary Table~\ref{tab:cohorts} gives the expanded map.}
\label{tab:cohort_main}
\begin{tabularx}{\linewidth}{@{}lrX@{}}
\toprule
Set & Papers & Inputs or role \\
\midrule
Study 2 selected & 200 & Starting originals for single edits \\
With admitted edits & 199 & 3,323 single-edit versions \\
Edit-ranking analysis & 198 & 3,321 versions meeting the eligibility rule \\
Study 3 search & 128 & Same papers under adaptive and random search \\
Standardization & 328 & The 200 Study 2 and 128 Study 3 originals combined \\
\bottomrule
\end{tabularx}
\end{table}

We fix every evaluation text and its order before target scoring (Figure~\ref{fig:pipeline}). The scoring program rejects changes to these saved versions, and neither target guides generation, checking, or search. The term \emph{sealed} refers to edited-input evaluation, not complete prior ignorance of the targets. Both targets had scored the original release texts earlier. Those scores were excluded from sampling, prompt development, edit selection, and search. The protocol was specified locally before edited-input scoring, without public preregistration.

\subsection{How are edits produced and checked?}
\label{sec:interventions}
GPT-6-Luna at maximum reasoning effort is asked for up to 24 paragraph rewrites per paper, each a version that reads more convincingly to reviewers. Each replaces one paragraph and receives a descriptive type label: clarity, flow, structure, emphasis, or framing; rhetorical emphasis may change. The writer must preserve facts, claim strength, attribution, numbers, equations, citations, headings, and figure/table references; it may neither add evidence nor move material between paragraphs. These are instructions, not guarantees or an independently validated taxonomy. Supplementary Section~\ref{app:prompts} gives the prompts, and Section~\ref{app:examples} shows two complete original/edited examples selected by a fixed rule.

Study 2 also requests an evidence removal and a length-matched deletion of non-result text, called a sham. The removal deletes whole paragraphs reporting one result that supports a quoted abstract or Introduction claim. The sham deletes background or related work with no results or claims of the paper, matched to within 20\% of the removal's length. Because coverage of prior work is a reviewing criterion, the sham may itself carry a legitimate acceptance cost. This tests the implemented deletion procedure, not scientific understanding in general.

An edit is \emph{admitted} only if it passes two checks. An invariant check protects numbers, equations, citations, cross-references, headings, and table blocks. A semantic check by GPT-6.1-Sol at high reasoning effort judges whether meaning, claim strength, scope, and attribution remain unchanged. It sees the original and edited text without scorer outputs, and passes an item only when the preservation conditions are judged satisfied. For deletions, only the declared span may disappear, and the remaining paper must stay coherent. Claude Opus 5.5 rechecks a seeded 10\% of calls, also without scores. Only this sampled recheck crosses providers; the writer and primary checker share one. Supplementary Section~\ref{app:prompts} gives the rules, and Section~\ref{sec:validity} measures their limitations.

Study~3 uses a one-pass polish by the same writer as a separate control. It may change any paragraph; the same checks are applied to each changed paragraph, and failures revert to the original.

\subsection{How does search select revisions?}
\label{sec:search}
Study 3 compares two ways to select combinations of admitted edits. Both use only \Surr's scores and allow at most eight edits, one per paragraph, with total character count changing by at most 5\% in either direction. Adaptive search adds the best improving edit that respects those limits at each step; random search tries sampled combinations and keeps the best according to \Surr. Because both policies select with \Surr, this contrast tests a search policy rather than surrogate guidance against unguided editing. Supplementary Section~\ref{app:search_protocol} gives the full algorithms and accounting rules.

We run two seeds per paper and record choices at maximum allowances of $Q\in\{1,8,64,512\}$ surrogate evaluations. Adaptive search can stop early, so the policies need not use the same number of evaluations. The chosen text is called an endpoint. At 64 and 512, the same primary semantic checker judges every nonempty endpoint as a complete document, using the preservation criteria above; failed endpoints contribute zero gain. Endpoints at one and eight lack this final check and are exploratory. The filtered one-pass polish is a separate control.

\subsection{How are agreement and search gains measured?}
\label{sec:statistics}
For paper $i$ and edit $e$, let $\Delta a_{ie}$ be the edited score minus the original score from evaluator $a$.
Within each paper, we rank edits by each scorer's score change and correlate the two rankings using Spearman's coefficient \citep{spearman1904}. Edit fidelity averages these correlations, giving every eligible paper equal weight:
\begin{equation}
\rho_T=\frac{1}{n}\sum_{i=1}^{n}\operatorname{Spearman}\!\left(\{\Delta S_{ie}\}_{e\in E_i},\{\Delta T_{ie}\}_{e\in E_i}\right).
\label{eq:fidelity}
\end{equation}
Here $S$ is the surrogate, $T$ is either target, $n$ is the number of eligible papers, and $E_i$ contains paper $i$'s admitted single edits. Papers need at least five edits and nonconstant changes for both compared evaluators.

For search, the comparison is adaptive gain minus random gain. We compute this difference for each seed, average the two seeds within a paper, then average the papers. Invalid or unedited choices contribute zero gain, so a failed check never removes a paper from the comparison.

To compare score changes, we divide by each evaluator's sample standard deviation (SD) across the 328 original papers, the 200 Study~2 and 128 Study~3 originals combined. A change of one SD is as large as that evaluator's typical between-paper variation, not a measure of scientific quality; it equals 0.153, 0.227, and 0.273 in acceptance score for \Surr, \Tone, and \Ttwo. We also give absolute changes in points, 100 times the raw acceptance score (Supplementary Section~\ref{app:estimands}).

We use 10,000 paper-level percentile bootstrap \citep{efron1993} resamples for the principal means, keeping edits and search seeds within paper. For each target, we test two outcomes: agreement on individual edits and adaptive-minus-random search gain at $Q=512$. These make four pre-specified tests (Table~\ref{tab:confirmatory}), run with the bfloat16 implementation. Holm's procedure at 0.05 \citep{holm} adjusts only these four two-sided bootstrap tail probabilities, controlling the chance of any false positive in that family.

Because the bfloat16 edit changes are largely rounding error (Section~\ref{sec:fidelity}), interpretation of the released weights uses the post hoc 32-bit rows of Table~\ref{tab:confirmatory}. These rows are outside the four-test family. 

Displayed 95\% confidence intervals (CIs) are nominal bootstrap intervals for individual estimates. They are not simultaneous intervals and are not adjusted for multiple comparisons. Guidance on significance testing and statistical power informs our interpretation \citep{hitchhiker,littlepower}. A result compatible with zero is not treated as evidence of equivalence.

Post hoc analyses, including the \Tone--\Ttwo{} comparison with Equation~\ref{eq:fidelity}, are outside the four-test family (Supplementary Section~\ref{app:posthoc}).

\section{How often are the proposed edits valid?}
\label{sec:validity}
Preserving numbers does not ensure unchanged claims. None of the 28 whole-paper rewrites generated in the pilot, under two prompt formats, passed the checks: failures changed hedges, novelty qualifications, or intended versus completed achievements. We therefore check paragraph edits and polish paragraph by paragraph.

Passing the automated checks is only the first test of preservation (Supplementary Figure~\ref{fig:validity}a). The two checks admit 70.5\% of checkable Study 2 paragraph edits but only 102 of 198 evidence removals, because the semantic checker rejects deletions that remove more than the named result or leave the paper incoherent. A second provider agrees with the primary checker on 86.6\% of 1,323 rechecked items, a measure of consistency rather than correctness. Supplementary Section~\ref{app:validity} gives the other pass rates, including those of search endpoints, and a planted-corruption challenge; no check estimates the miss rate for natural edits.

\paragraph{Blind human audit after scoring}
The blind audit asks whether items already admitted by the checks are valid on human inspection. One author judges 99 items while blind to evaluator scores, checker verdicts, and search policy: 59 endpoints, 20 Study 2 presentation edits, and 20 evidence removals. The endpoint sample contains one endpoint per selected paper; Supplementary Section~\ref{app:validity} describes the sampling and judgment procedure. All 59 endpoints and 20 presentation edits are judged valid. Exact one-sided 95\% upper bounds on invalidity are 4.95\% and 13.9\%, respectively. The endpoint bound pools the search policies (counts by policy in Supplementary Table~\ref{tab:human}).

The audit judges six of 20 removals invalid: 30.0\%, with a two-sided exact 95\% CI of 11.9--54.3\%. All six leave a coherence problem, such as a dangling connective, an experiment count that no longer fits, or an empty heading. Two also remove more than one result. The admitted removal intervention therefore mixes missing evidence with unintended coherence changes. We interpret it as removal performed by this editing procedure, rather than as a clean ablation of scientific evidence.

The audit does not filter the primary analysis. The audit has one rater, who is an author; there is no independent inter-rater reliability estimate. Undetected content changes could raise or lower agreement, depending on how they reorder each scorer's responses, and the audit cannot tell which.

\paragraph{Adjudicating checker disagreements}
A separate task examines disagreements between checkers. The same author judges 30 items accepted by the primary checker but rejected by the second provider: 13 are invalid and one is uncertain. One item overlaps with the blind audit, leaving 29 additional items. These selected disagreements show real checker misses, but cannot estimate errors shared by both checkers. Supplementary Section~\ref{app:validity} gives the audit overlap, review procedure, failure categories, and overlap with the invariant check. The primary analysis remains unchanged; passing the automated checks is not a guarantee of content preservation.

\begin{table}[!b]
\centering
\caption{Pre-specified edit-fidelity and search tests. Edit fidelity is the mean within-paper Spearman correlation \citep{spearman1904}; the search contrast is adaptive minus random target gain at a maximum budget of 512 surrogate evaluations, in target standard deviation (SD) units, with both policies selecting by the surrogate. Confidence intervals (CIs) are nominal 95\% paper-bootstrap intervals; $p$-values are Holm-adjusted across the four pre-specified rows \citep{holm}, which use the bfloat16 implementation (Section~\ref{sec:methods}). The rows after the four pre-specified tests are post hoc, descriptive, uncorrected for multiplicity, and outside the testing family: the edit-fidelity estimand applied to the two targets, and edit fidelity, search contrast, and target agreement with the targets rescored in 32-bit floating point.}
\label{tab:confirmatory}
\begin{tabular}{llrcc}
\toprule
Test & Target & Papers & Estimate [95\% CI] & Adjusted $p$ \\
\midrule
Edit fidelity & \Tone & 198 & \CI{-0.008}{-0.041}{0.025} & 1.0 \\
 & \Ttwo & 198 & \CI{-0.015}{-0.052}{0.022} & 1.0 \\
Search contrast & \Tone & 128 & \CI{0.0021}{-0.0190}{0.0240} & 1.0 \\
 & \Ttwo & 128 & \CI{-0.0005}{-0.0136}{0.0124} & 1.0 \\
\midrule
Target agreement & \Tone{} vs \Ttwo & 198 & \CI{0.033}{-0.008}{0.075} & -- \\
\multicolumn{5}{l}{\emph{Targets rescored in 32-bit floating point (post hoc; estimates for the released weights)}} \\
Edit fidelity & \Tone & 198 & \CI{-0.008}{-0.047}{0.031} & -- \\
 & \Ttwo & 198 & \CI{-0.021}{-0.059}{0.016} & -- \\
Search contrast & \Tone & 128 & \CI{-0.0152}{-0.0328}{0.0015} & -- \\
 & \Ttwo & 128 & \CI{-0.0060}{-0.0218}{0.0089} & -- \\
Target agreement & \Tone{} vs \Ttwo & 198 & \CI{0.047}{0.004}{0.090} & -- \\
\bottomrule
\end{tabular}
\end{table}

\section{Results}
\label{sec:results}
\begin{figure}[t]
\centering
\includegraphics[width=\linewidth]{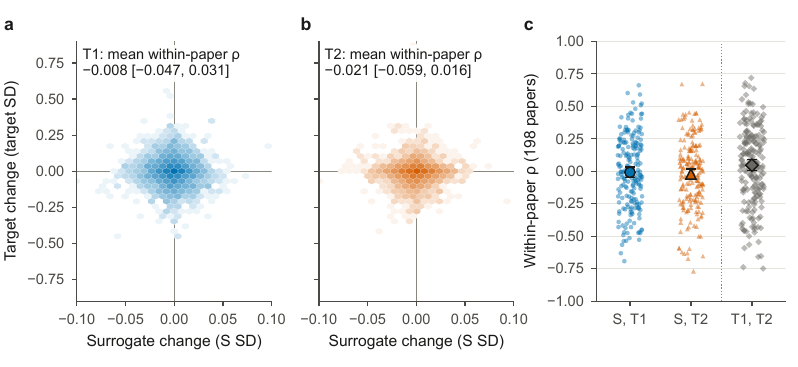}
\caption{Surrogate changes barely track target changes for the same edits. (a, b)~Density of all 3,323 admitted single edits across 199 papers: surrogate change against the change for \Tone{} (a) and \Ttwo{} (b) in the post hoc 32-bit rescoring of the released weights, each in units of its evaluator's standard deviation (SD) across unedited papers (the 328 frozen originals for \Surr; the 200 Study~2 originals, in 32-bit, for the targets); darker bins hold more edits (log scale). The axes use different standardized ranges because the surrogate moves far less. The inset gives the mean within-paper Spearman correlation \citep{spearman1904} with its 95\% paper-bootstrap confidence interval (CI); Table~\ref{tab:confirmatory} gives these 32-bit estimates and the pre-specified bfloat16 ones. (c)~Within-paper correlations for the 198 eligible papers between \Surr{} and each target and, right of the dotted rule, between the two targets (post hoc, 32-bit; points, offset horizontally so equal values stay visible), with their mean and 95\% CI (outlined marker).}
\label{fig:fidelity}
\end{figure}

\subsection{Original-paper agreement is not detected on edits}
\label{sec:fidelity}
In the pre-specified bfloat16 analysis, the classifier shows no detectable edit-ranking agreement with either target. Across 198 papers with at least five admitted edits and varying scores, mean within-paper correlations are $-0.008$ for \Tone{} and $-0.015$ for \Ttwo{} (Table~\ref{tab:confirmatory}). Their nominal 95\% upper limits are 0.025 and 0.022, leaving little support for positive mean agreement in this edit population. These bounds concern differences among edits within a paper, whereas original-paper correlations concern differences between whole papers. Section~\ref{sec:headroom} therefore also measures the practical score gain from selecting edits.

On the 198 papers used here rather than the 1,667-paper benchmark, \Surr's rank correlation is 0.724 with \Tone{} and 0.479 with \Ttwo, so the paper-versus-edit contrast is not a cohort difference (Supplementary Section~\ref{app:extended}).

The null is not confined to the sparse classifier. A post hoc dense surrogate that reads word order also shows no edit agreement with the bfloat16 targets. Copies of \Surr{} trained with the same recipe on disjoint halves of its data agree only weakly with each other on edits. A regularized linear predictor fitted to target changes finds no agreement on held-out papers, but also recovers \Surr's own changes poorly, limiting that check. Supplementary Section~\ref{app:postreview} gives these models and estimates.

The targets themselves agree weakly on edits despite agreeing on original papers. On the same 198 papers, their original-paper correlation is 0.603, while their post hoc 32-bit edit agreement is \CI{0.047}{0.004}{0.090} (Table~\ref{tab:confirmatory}; Figure~\ref{fig:fidelity}c). They share an architecture and base checkpoint but differ in training collection and prompt. Swapping \Tone's prompt for \Ttwo's changes its edit rankings much more than its paper rankings and does not close the target-agreement gap (Supplementary Section~\ref{app:reliability}).

\paragraph{Response size and numerical precision}
Numerical precision matters because the rewrites cause small score changes. At full precision, mean absolute responses over all 199 papers with admitted edits range from 1.0 to 1.4 raw-score points out of 100. The pre-specified bfloat16 implementation produces larger changes dominated by rounding, and its edit changes agree only weakly with the 32-bit ones (Supplementary Section~\ref{app:reliability}).

A hardware check separates this precision effect from reproducibility. Scoring in 32-bit on a second GPU reproduces edit changes almost exactly (0.9997 and 0.9998 on 25 papers; Supplementary Section~\ref{app:reliability}). With 32-bit targets, the edit-fidelity estimates remain near zero (Table~\ref{tab:confirmatory}).

\paragraph{Stability under whitespace changes}
A whitespace-only control measures sensitivity to a change that leaves the wording intact. On a matched 60-paper subset, adding a space moves the 32-bit target scores about half as much as an ordinary rewrite (Supplementary Section~\ref{app:reliability}).

To test whether this sensitivity disrupts edit rankings, we rescore the rewrites from a seeded 60-paper subset after adding one space between two sentences away from the edited span. For each target, we correlate its edit-score changes before and after adding the space. The mean within-paper correlations are \CI{0.760}{0.713}{0.804} for \Tone{} and \CI{0.743}{0.692}{0.783} for \Ttwo{} in 32-bit.

These rankings are fairly stable under this one perturbation, but stability does not establish sensitivity to scientific content. Other perturbations could reveal lower reliability. Supplementary Section~\ref{app:reliability} gives the conditional attenuation comparison.

\paragraph{Larger responses and evidence removal}
The narrow agreement bound covers the full eligible edit set. To check the subset that moves a target more than the whitespace control, we retain responses above each target's own control-response 95th percentile: agreement with \Surr{} is \CI{0.050}{-0.068}{0.166} for \Tone{} and \CI{-0.045}{-0.166}{0.078} for \Ttwo{} (62 and 50 papers), so the study cannot bound agreement tightly on the edits the targets respond to most.

Evidence removal asks a different question: whether the targets distinguish lost results from a matched deletion of other text. The paired result-removal and sham deletions have no detectable difference in either target's score, in bfloat16 or in 32-bit (93 papers; Supplementary Section~\ref{app:removal}). However, audited removals often damaged coherence (Section~\ref{sec:validity}). The intervals also allow content responses as large as a mean edit response. This test therefore does not show that the targets ignore content.

\begin{figure}[t]
\centering
\includegraphics[width=\linewidth]{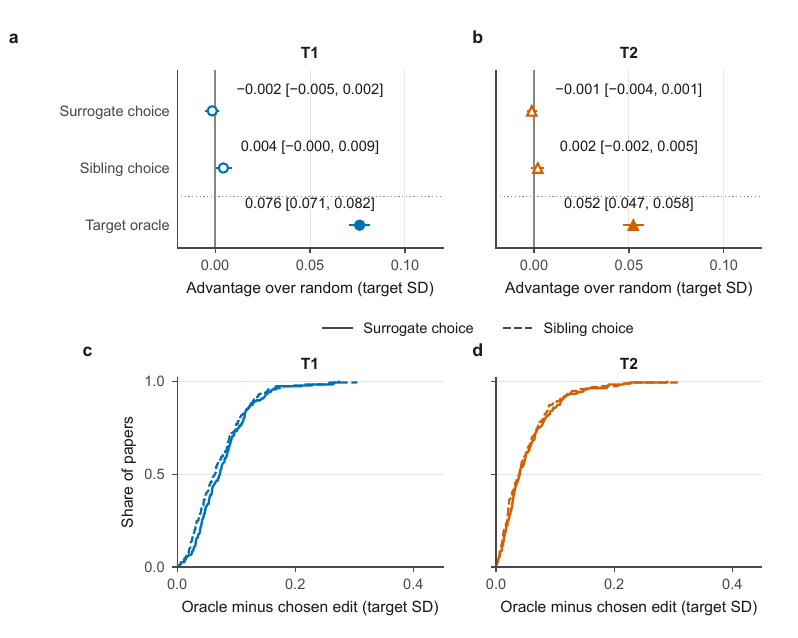}
\caption{Proxy choices give little average advantage in equal-size five-edit banks (post hoc). (a, b)~Chosen target change minus the exact uniform-random mean of the same sampled bank, for surrogate choice, sibling choice, and the evaluated target's oracle choice. Open markers denote proxy choices; filled markers denote the outcome-informed oracle. Estimates average 1,000 draws within each of 198 papers; intervals are nominal 95\% paper-bootstrap confidence intervals. Target changes are the 32-bit rescoring of the released weights, in units of each target's 32-bit standard deviation (SD) across the 200 Study~2 originals; bfloat16 values are in Supplementary Table~\ref{tab:fixed_bank}. (c, d)~Empirical cumulative distributions of paperwise oracle-minus-choice regret, averaged over those same draws; solid lines use the surrogate, dashed lines the sibling. The intervals are conditional on the existing edit banks and exclude Monte Carlo subsampling error, which Supplementary Section~\ref{app:posthoc} quantifies.}
\label{fig:headroom}
\end{figure}

\paragraph{Controls that produce agreement}
The cross-GPU check establishes reproducible rankings, but does the estimand register agreement between different scorers? In a bfloat16 positive control, graded paragraph deletions produce agreement across all three pairs; length-preserving word shuffles produce agreement only for \Surr--\Tone. The controls damage progressively larger sets of paragraphs, with each severe version containing the milder damage. This gives an easier graded signal than equal-size edits (Supplementary Section~\ref{app:postreview}).

A further control removes the nesting as well as the length change. For each of 50 papers, we create ten versions, each scrambling a different single paragraph. The versions keep their length and are not nested. Using float16 as a proxy for the release recipe, the two targets agree at \CI{0.226}{0.128}{0.316}, whereas \Surr{} agrees with neither (\CI{-0.068}{-0.144}{0.010} and \CI{-0.043}{-0.141}{0.057}; Supplementary Section~\ref{app:postreview}); the estimand thus registers agreement between the targets on equal-size changes, although no tested equal-size change set shows agreement between \Surr{} and a target.

\subsection{Adaptive search raises the surrogate little and the targets not detectably}
\label{sec:transfer}
At the pre-specified maximum budget of 512, adaptive search does not measurably outperform random search on either target, in bfloat16 or in 32-bit (Table~\ref{tab:confirmatory}); other budgets are exploratory (Supplementary Figure~\ref{fig:transfer}). The search barely raises its own objective: at 512, its mean increase over the unedited paper is 0.064 surrogate SD, versus 0.053 for random search, a difference of \CI{0.011}{0.008}{0.013} SD. A target gain of that size lies inside both bfloat16 intervals. In 32-bit, however, it exceeds both upper limits (0.0015 and 0.0089 SD). Thus, a gain as large as the surrogate's is excluded on each target's own SD scale; equal SD units need not mean equal transfer. Neither policy's target gain over the original excludes zero (Supplementary Table~\ref{tab:changes}); twelve adaptive and ten random endpoints fail the final semantic-preservation check and remain in the analysis with zero gain.

\subsection{Fixed-size edit banks reveal little proxy-selection advantage}
\label{sec:headroom}
This post hoc check compares choices from the same five-edit bank against its average target gain, the expected result of uniform random choice. The sibling chooser is \Ttwo{} when evaluating \Tone{}, and conversely. Choosing with the evaluated target supplies an outcome-informed oracle, which requires target access.

Neither \Surr{} nor the sibling model shows a measurable average advantage over random choice (Figure~\ref{fig:headroom}a, b). By contrast, choosing with the evaluated target gives advantages of \CI{0.076}{0.071}{0.082} SD for \Tone{} and \CI{0.052}{0.047}{0.058} for \Ttwo{}. These are 1.8 and 1.3 raw-score points, more than twice the mean change from an added space. Surrogate and sibling choices stay within 0.005 SD of random choice. This positive advantage follows from selecting the highest target score; it measures available score variation, not semantic improvement. The bfloat16 estimates are in Supplementary Table~\ref{tab:fixed_bank}.

Larger candidate sets increase the oracle's advantage, but not the proxies' measurable advantage, on a fixed cohort (Supplementary Figure~\ref{fig:bank_size}). The finding concerns small presentation-level responses; it does not show that edit preferences diverge where responses are large.

{\let\FloatBarrier\relax\FloatBarrier
\section{Related work}
\label{sec:related}
\paragraph{Acceptance prediction and revision feedback}
Predicting acceptance from paper text predates large language models: PeerRead paired submissions with their decisions and trained acceptance classifiers on the text \citep{peerread}. Natural language processing (NLP) work on peer review spans generated reviews \citep{yuanreview}, shared datasets, and assistance tasks \citep{nlpeer,nlp4review}. LLM feedback on research papers overlaps substantially with points raised by human reviewers \citep{liangfeedback}, while evaluations of LLM reviewers report clear limits in scoring and critique \citep{reliablereviewer,critiquereview}. IteraTeR annotates iterative edits and their intentions across several domains \citep{iterater}; arXivEdits aligns scientific paper versions and annotates edit intentions \citep{arxivedits}. ARIES links paper edits to reviewer comments \citep{aries}. These resources mix content and presentation changes and do not cover the held-out ICLR 2026 papers, so we use checked synthetic rewrites. 

Table~\ref{tab:prior_comparisons} distinguishes the closest comparisons from our question. PaperScorer's feature baseline reports 61.1\% accuracy, a different metric from our AUC. Its section-edit gains of 3.9--6.1\% also differ in unit from our raw-score points. We did not rerun those revisions or assess the size or validity of its gains. Supplementary Section~\ref{app:postreview} compares the between-paper and within-paper estimands.

\begin{table}[tb]
\centering
\small
\caption{The closest comparisons address different questions and units of analysis.}
\label{tab:prior_comparisons}
\begin{tabularx}{\linewidth}{@{}p{0.20\linewidth}p{0.36\linewidth}X@{}}
\toprule
Study & Outcome examined & Relation to this study \\
\midrule
PaperScorer \citep{paperscorer} & Own-model feedback on section rewrites of 30 borderline papers & We select revisions with a separate classifier. \\
\citet{gainagreement} & Across-paper Pearson correlations: 0.804 on original scores; 0.085--0.415 on revision gains & We rank several paragraph alternatives within each paper. \\
RobustReview \citep{robustreview} & Stability under preserving rewrites and discrimination across papers & These are complementary to agreement between scorers on edits. \\
This study & Within-paper edit rankings, search, and choices from equal-size banks & Edits are fixed before target scoring; preservation errors limit interpretation. \\
\bottomrule
\end{tabularx}
\end{table}

\paragraph{Manipulation and transfer}
Substitute models can generate adversarial inputs that transfer to a target whose internals are unknown \citep{blackbox,whytransfer}, including word-level attacks on text classifiers \citep{textfooler}. Transfer depends partly on how well the two models' input gradients align \citep{whytransfer}. Our edit-fidelity measure compares responses to a discrete set of edits. Interpretability uses surrogates to explain individual predictions \citep{lime}, and model extraction asks how far a substitute reproduces a target's outputs \citep{tramer2016}. 

Other attacks manipulate the scored input directly. Distracting sentences and input-agnostic trigger sequences alter NLP outputs \citep{jialiang,triggers}; TextAttack provides standardized search methods for such perturbations \citep{textattack}. Optimizing a learned proxy can raise its score while the objective it stands for stalls or declines \citep{rewardoveropt}, a failure formalized as reward hacking \citep{rewardhacking} and classified under Goodhart's law \citep{goodhart}. 

In peer review, repackaging raises review scores and can transfer to unqueried reviewers \citep{repackaging}; abstract rephrasing also transfers without knowing the reviewing model \citep{abstractattack}. PaperGuard studies surrogate-based attacks on multimodal review \citep{paperguard}. Prompt injection instead embeds explicit instructions in papers \citep{injection}, unlike our paragraph rewrites. Earlier surrogate-based modifications targeted unknown paper--reviewer assignment systems \citep{assignment}. 

Predictive accuracy and response to optimization are distinct. Equally accurate classifiers can conflict on individual predictions \citep{marx2020}, and transfer attacks can exploit shared adversarial directions and nearby decision boundaries \citep{transferspace}. Reward models with similar in-distribution accuracy can diverge once a policy is optimized against them; ensembling reduces overoptimization \citep{coste2024,eisenstein2024} but does not remove reward hacking \citep{eisenstein2024}. Essays written to exploit an automated scorer can receive scores unlike human raters' \citep{powers2002}, and short-answer scorers can be fooled by answers with shuffled words \citep{ding2020}. Our result measures this known gap for released acceptance predictors on checker-admitted paragraph rewrites; the contribution is the measurement, not the phenomenon. Here transfer means that edits selected with one scorer also raise another scorer's output.

\paragraph{Scientific content, rhetoric, and validation}
Counterfactual evaluation probes whether automatic reviewers detect faulty research logic \citep{counterfactual}. Controlled rhetorical rewrites reveal structured changes in review judgments \citep{rhetoric}, while RobustReview treats stability under preserving rewrites and discrimination across papers as a joint requirement (Table~\ref{tab:prior_comparisons}). We compare response magnitudes and edit-level agreement, rather than proposing a robust reviewer. Testing models on minimally edited inputs, rather than only on held-out accuracy, follows behavioral testing and contrast sets \citep{checklist,contrastsets}; counterfactually augmented data also use revised examples for training \citep{cad}. 

LLM evaluators can correlate with human ratings \citep{geval} yet shift judgments with candidate order \citep{unfairevaluators,koobias}, familiar text or small prompt changes \citep{inconsistent}, and their own generations \citep{selfpref}. 

Evaluating revisions also requires knowing what changed. This is needed to check whether a revision addresses a reviewer's concern \citep{autosupervision} or to separate style from content \citep{rhetoricalstyle}. Our preservation checks serve this role, and their errors limit the agreement measurement.
}
\section{Discussion}
\label{sec:discussion}
\subsection{What original-paper agreement leaves unmeasured}
The surrogate and the two acceptance models agree more on existing papers than on revisions. The surrogate closely tracks the first target's original-paper rankings, yet barely tracks which rewrites raise its score. The two targets also share some response to edits, but their agreement is weak. This distinction matters whenever one scorer stands in for another during revision selection.

Paper prediction and edit response can depend on different textual properties. Our checks hold most document-structure counts fixed, so the edit comparison mainly probes word features and length. Those word features also carry much of the surrogate's original-paper agreement with the first target; holding structure fixed alone therefore does not explain that pair's gap. For the second target, original-paper agreement depends more on structure, making it a weaker premise for revision selection. Supplementary Section~\ref{app:postreview} gives these feature comparisons and a geometric description of the directions the edits change. They motivate possible accounts, rather than establish a mechanism.

Acceptance labels are noisy \citep{cortes2021,beygelzimer2023}. They report paper-level decisions, rather than preferences among presentation rewrites. Copies of the surrogate trained on disjoint halves of its data give different edit rankings, consistent with such preferences being weakly constrained by those labels. Yet the targets share some response magnitudes after accounting for edit size, position, and type, and they agree on rankings of equal-size paragraph scrambles. Thus these change sets contain shared signal, even though the surrogate does not capture the targets' edit rankings. Input sensitivity could still dominate the targets' responses; the analyses do not identify a causal mechanism inside either model.

Different input lengths remain a limit. Every admitted changed paragraph is fully visible to both targets, while some extend beyond the surrogate's shorter word window. Restricting the analysis to paragraphs visible to all scorers leaves agreement near zero (Supplementary Table~\ref{tab:visibility} and Figure~\ref{fig:visibility}). This still leaves different surrounding text and whole-document features, and a length change can shift a target's truncation boundary.

\subsection{Scope and limitations}
The surrogate scenario motivates the measurement but is not a documented practice, and the recommendation covers checker-admitted presentation-only paragraph rewrites, not content-bearing or section-level revisions. The evidence covers one sparse surrogate, a post hoc dense surrogate, one writer configuration, English ICLR papers, and two acceptance predictors trained from the same base language model.

We also test DeepReviewer-14B, a generative reviewer from another model family \citep{deepreviewer}. The score comes from one forced rating in an 8-bit version whose outputs were not validated against the original weights. It shows no edit-ranking agreement with the surrogate or either target, but also fails the equal-size scramble control. This check therefore cannot establish that the main null extends to other model families (Supplementary Section~\ref{app:deepreviewer}).

Several extensions remain untested: other editors, content-changing and human revisions, and broader interventions. We also leave open nonlinear surrogates trained on edit outcomes or target scores, theory-derived rhetorical measures, such as signposting and reader address, and qualitative analysis of the largest responses. Model coverage does not include PaperScorer's page-image and larger models, other generative and closed reviewers, or equalized model inputs. Earlier transfer results came from presentation-only revisions of the paper \citep{repackaging} and adversarial rewrites of the abstract \citep{abstractattack}, optimized against reviewers that read the same text; our checked paragraph edits and the surrogate's different input could each reduce transfer. 

Preservation is also uncertain. The single-author review of checker disagreements exposed missed content changes (Section~\ref{sec:validity}). No human panel assessed paper quality or acceptance prospects, so score movement is not evidence of scientific merit; a planned human-reviewer comparison was dropped for lack of version-matched histories.

The target scores themselves limit interpretation (Section~\ref{sec:fidelity}). The release's own recipe, which serves the weights in float16 through vLLM, was not run. A float16 pass in Transformers reproduces the 32-bit edit changes and benchmark AUCs and leaves edit fidelity near zero. In 32-bit, whitespace-only changes still move both targets in this implementation, which passes raw text. We did not check whether the release's preprocessing normalizes whitespace. The design cannot tell whether the response reflects wording that the surrogates miss or sensitivity to formatting. 

The design was specified locally without public preregistration, the human audit followed scoring contrary to the earlier plan, and the post hoc analyses followed the four pre-specified tests.

The practical consequence is to test revision tools on the changes they are meant to guide. The same checks could be applied to LLM judges and learned reward models when a cheaper scorer selects texts for another evaluator. Our study motivates those tests; it does not show that these other systems fail them.

\section{Conclusion}
In two released acceptance predictors, agreement on existing papers was not matched by detectable agreement on checked paragraph rewrites. The classifier tracks original-paper rankings closely for one target and moderately for the other, yet tracks neither's edit rankings. The targets also barely agree on edits. Choosing edits with the classifier gives no detectable advantage over random choice. Even choosing the best of five with the predictor itself raises its score by under two points in 32-bit, more than twice the effect of an added space (Section~\ref{sec:headroom}). These results bound agreement for checker-admitted rewrites rather than show that either predictor is wrong. Revision tools should therefore be validated against their intended evaluators on the edits they select, with response reliability reported. Agreement between models does not establish valid edit preferences.

\section*{Broader impact statement}
Automated paper scores can shape authors' effort and research assessment. In this study, a cheap classifier's agreement with acceptance predictors on existing papers did not indicate which revisions those predictors reward, and a constrained search that raised the classifier's own score by about 0.06 SD did not detectably raise theirs. These observations support testing feedback tools on controlled revisions and not relying on their scores alone for publication decisions. Apart from this article and two licensed example paragraphs, the planned release contains no manuscript text, and it contains no author-facing submission optimizer.

\section*{Data and code availability}
The experiments use the publicly released PaperScorer text dataset and checkpoints \citep{paperscorer} and the DeepReviewer-14B checkpoint \citep{deepreviewer}. The analysis code, paper identifiers, edit types and positions, scores, non-text metadata, and experiment logs are released at \url{https://github.com/nbbllxx0/Ranking-Agreements}, which is made public once the arXiv preprint of this article is available. Apart from the two licensed examples in Supplementary Section~\ref{app:examples}, original and edited manuscript text will not be redistributed publicly; the frozen inputs, including the edited texts, are available to the editor and reviewers on request for verification only, under the terms of the source release. The released per-edit scores reproduce Table~\ref{tab:confirmatory} and the agreement estimates without the texts; refitting models on the texts or computing text descriptors needs the frozen inputs. PaperScorer's dataset and model cards state custom or ``other'' licensing, and the base models have their own terms; we do not treat them as freely redistributable. DeepReviewer use is subject to its model license and its prohibition on official peer reviewing and publication decisions.

\section*{Acknowledgments}
We used language models in two separate ways. As research instruments, described in the paper, GPT-6-Luna rewrote paragraphs of the ICLR papers to create the experimental edits, GPT-6.1-Sol checked them, and Claude Opus~5.5 rechecked a sample; DeepReviewer-14B, in quantized form, was used only as an evaluation target in this non-commercial robustness study, and its outputs were not used as peer reviews or to make publication decisions. Separately, two artificial intelligence (AI) assistants, Claude Opus~5.5 and GPT-6.1-Sol, helped us implement the analysis and plotting scripts, edit the text of this article, and check its numbers and references against the analysis outputs and the cited sources. The authors set the research objective and the protocol decisions, performed the blind human audit and the adjudication of checker disagreements, wrote the manuscript, reviewed and revised all text and code, and take full responsibility for the content of the article and its conclusions. No AI system is an author.
\FloatBarrier
\bibliographystyle{plainnat}
\bibliography{references}

\begin{thebibliography}{74}
\providecommand{\natexlab}[1]{#1}
\providecommand{\url}[1]{\texttt{#1}}
\expandafter\ifx\csname urlstyle\endcsname\relax
  \providecommand{\doi}[1]{doi: #1}\else
  \providecommand{\doi}{doi: \begingroup \urlstyle{rm}\Url}\fi

\bibitem[Abdin et~al.(2024)Abdin, Aneja, Behl, Bubeck, Eldan, Gunasekar,
  et~al.]{phi4}
Marah Abdin, Jyoti Aneja, Harkirat Behl, S{\'e}bastien Bubeck, Ronen Eldan,
  Suriya Gunasekar, et~al.
\newblock {Phi-4} technical report.
\newblock \emph{arXiv preprint arXiv:2412.08905}, 2024.
\newblock URL \url{https://arxiv.org/abs/2412.08905}.

\bibitem[Beygelzimer et~al.(2023)Beygelzimer, Dauphin, Liang, and
  Wortman~Vaughan]{beygelzimer2023}
Alina Beygelzimer, Yann~N. Dauphin, Percy Liang, and Jennifer Wortman~Vaughan.
\newblock Has the machine learning review process become more arbitrary as the
  field has grown? {The} {NeurIPS} 2021 consistency experiment.
\newblock \emph{arXiv preprint arXiv:2306.03262}, 2023.

\bibitem[Card et~al.(2020)Card, Henderson, Khandelwal, Jia, Mahowald, and
  Jurafsky]{littlepower}
Dallas Card, Peter Henderson, Urvashi Khandelwal, Robin Jia, Kyle Mahowald, and
  Dan Jurafsky.
\newblock With little power comes great responsibility.
\newblock In \emph{Proceedings of the 2020 Conference on Empirical Methods in
  Natural Language Processing (EMNLP)}, pages 9263--9274. Association for
  Computational Linguistics, 2020.
\newblock \doi{10.18653/v1/2020.emnlp-main.745}.

\bibitem[Clopper and Pearson(1934)]{clopperpearson}
C.~J. Clopper and E.~S. Pearson.
\newblock The use of confidence or fiducial limits illustrated in the case of
  the binomial.
\newblock \emph{Biometrika}, 26\penalty0 (4):\penalty0 404--413, 1934.
\newblock \doi{10.1093/biomet/26.4.404}.

\bibitem[Cortes and Lawrence(2021)]{cortes2021}
Corinna Cortes and Neil~D. Lawrence.
\newblock Inconsistency in conference peer review: Revisiting the 2014
  {NeurIPS} experiment.
\newblock \emph{arXiv preprint arXiv:2109.09774}, 2021.

\bibitem[Coste et~al.(2024)Coste, Anwar, Kirk, and Krueger]{coste2024}
Thomas Coste, Usman Anwar, Robert Kirk, and David Krueger.
\newblock Reward model ensembles help mitigate overoptimization.
\newblock In \emph{International Conference on Learning Representations
  (ICLR)}, 2024.
\newblock URL
  \url{https://proceedings.iclr.cc/paper_files/paper/2024/hash/dda7f9378a210c25e470e19304cce85d-Abstract-Conference.html}.

\bibitem[D'Amour et~al.(2022)D'Amour, Heller, Moldovan, et~al.]{damour2022}
Alexander D'Amour, Katherine Heller, Dan Moldovan, et~al.
\newblock Underspecification presents challenges for credibility in modern
  machine learning.
\newblock \emph{Journal of Machine Learning Research}, 23\penalty0
  (226):\penalty0 1--61, 2022.
\newblock URL \url{https://jmlr.org/papers/v23/20-1335.html}.

\bibitem[D'Arcy et~al.(2024)D'Arcy, Ross, Bransom, Kuehl, Bragg, Hope, and
  Downey]{aries}
Mike D'Arcy, Alexis Ross, Erin Bransom, Bailey Kuehl, Jonathan Bragg, Tom Hope,
  and Doug Downey.
\newblock {ARIES}: A corpus of scientific paper edits made in response to peer
  reviews.
\newblock In \emph{Proceedings of the 62nd Annual Meeting of the Association
  for Computational Linguistics (Volume 1: Long Papers)}, pages 6985--7001.
  Association for Computational Linguistics, 2024.
\newblock \doi{10.18653/v1/2024.acl-long.377}.

\bibitem[Demontis et~al.(2019)Demontis, Melis, Pintor, Jagielski, Biggio,
  Oprea, Nita-Rotaru, and Roli]{whytransfer}
Ambra Demontis, Marco Melis, Maura Pintor, Matthew Jagielski, Battista Biggio,
  Alina Oprea, Cristina Nita-Rotaru, and Fabio Roli.
\newblock Why do adversarial attacks transfer? {E}xplaining transferability of
  evasion and poisoning attacks.
\newblock In \emph{28th USENIX Security Symposium (USENIX Security 19)}, pages
  321--338, Santa Clara, CA, 2019. USENIX Association.
\newblock URL
  \url{https://www.usenix.org/conference/usenixsecurity19/presentation/demontis}.

\bibitem[Ding et~al.(2020)Ding, Riordan, Horbach, Cahill, and Zesch]{ding2020}
Yuning Ding, Brian Riordan, Andrea Horbach, Aoife Cahill, and Torsten Zesch.
\newblock Don't take ``nswvtnvakgxpm'' for an answer -- the surprising
  vulnerability of automatic content scoring systems to adversarial input.
\newblock In \emph{Proceedings of the 28th International Conference on
  Computational Linguistics}, pages 882--892, Barcelona, Spain (Online), 2020.
  International Committee on Computational Linguistics.
\newblock \doi{10.18653/v1/2020.coling-main.76}.

\bibitem[Dror et~al.(2018)Dror, Baumer, Shlomov, and Reichart]{hitchhiker}
Rotem Dror, Gili Baumer, Segev Shlomov, and Roi Reichart.
\newblock The hitchhiker's guide to testing statistical significance in natural
  language processing.
\newblock In \emph{Proceedings of the 56th Annual Meeting of the Association
  for Computational Linguistics (Volume 1: Long Papers)}, pages 1383--1392.
  Association for Computational Linguistics, 2018.
\newblock \doi{10.18653/v1/P18-1128}.

\bibitem[Du et~al.(2024)Du, Wang, Zhao, Deng, Liu, Lou, Zou, Narayanan~Venkit,
  Zhang, Srinath, Zhang, Gupta, Li, Li, Wang, Liu, Liu, Gao, Xia, Xing,
  Jiayang, Wang, Su, Shah, Guo, Gu, Li, Wei, Wang, Cheng, Ranathunga, Fang, Fu,
  Liu, Huang, Blanco, Cao, Zhang, Yu, and Yin]{critiquereview}
Jiangshu Du, Yibo Wang, Wenting Zhao, Zhongfen Deng, Shuaiqi Liu, Renze Lou,
  Henry~Peng Zou, Pranav Narayanan~Venkit, Nan Zhang, Mukund Srinath,
  Haoran~Ranran Zhang, Vipul Gupta, Yinghui Li, Tao Li, Fei Wang, Qin Liu,
  Tianlin Liu, Pengzhi Gao, Congying Xia, Chen Xing, Cheng Jiayang, Zhaowei
  Wang, Ying Su, Raj~Sanjay Shah, Ruohao Guo, Jing Gu, Haoran Li, Kangda Wei,
  Zihao Wang, Lu~Cheng, Surangika Ranathunga, Meng Fang, Jie Fu, Fei Liu,
  Ruihong Huang, Eduardo Blanco, Yixin Cao, Rui Zhang, Philip~S. Yu, and
  Wenpeng Yin.
\newblock {LLMs} assist {NLP} researchers: Critique paper (meta-)reviewing.
\newblock In \emph{Proceedings of the 2024 Conference on Empirical Methods in
  Natural Language Processing}, pages 5081--5099. Association for Computational
  Linguistics, 2024.
\newblock \doi{10.18653/v1/2024.emnlp-main.292}.

\bibitem[Du et~al.(2022)Du, Raheja, Kumar, Kim, Lopez, and Kang]{iterater}
Wanyu Du, Vipul Raheja, Dhruv Kumar, Zae~Myung Kim, Melissa Lopez, and Dongyeop
  Kang.
\newblock Understanding iterative revision from human-written text.
\newblock In \emph{Proceedings of the 60th Annual Meeting of the Association
  for Computational Linguistics (Volume 1: Long Papers)}, pages 3573--3590.
  Association for Computational Linguistics, 2022.
\newblock \doi{10.18653/v1/2022.acl-long.250}.

\bibitem[Dycke and Gurevych(2026)]{counterfactual}
Nils Dycke and Iryna Gurevych.
\newblock Automatic reviewers fail to detect faulty reasoning in research
  papers: A new counterfactual evaluation framework.
\newblock \emph{Transactions of the Association for Computational Linguistics},
  14:\penalty0 465--488, 2026.
\newblock \doi{10.1162/tacl.a.642}.
\newblock URL \url{https://aclanthology.org/2026.tacl-1.22/}.

\bibitem[Dycke et~al.(2023)Dycke, Kuznetsov, and Gurevych]{nlpeer}
Nils Dycke, Ilia Kuznetsov, and Iryna Gurevych.
\newblock {NLPeer}: A unified resource for the computational study of peer
  review.
\newblock In \emph{Proceedings of the 61st Annual Meeting of the Association
  for Computational Linguistics (Volume 1: Long Papers)}, pages 5049--5073.
  Association for Computational Linguistics, 2023.
\newblock \doi{10.18653/v1/2023.acl-long.277}.

\bibitem[Efron and Tibshirani(1993)]{efron1993}
Bradley Efron and Robert~J. Tibshirani.
\newblock \emph{An Introduction to the Bootstrap}.
\newblock Springer US, 1993.
\newblock \doi{10.1007/978-1-4899-4541-9}.

\bibitem[Eisenhofer et~al.(2023)Eisenhofer, Quiring, M{\"o}ller, Riepel, Holz,
  and Rieck]{assignment}
Thorsten Eisenhofer, Erwin Quiring, Jonas M{\"o}ller, Doreen Riepel, Thorsten
  Holz, and Konrad Rieck.
\newblock {No more Reviewer \#2: Subverting Automatic Paper-Reviewer Assignment
  using Adversarial Learning}.
\newblock In \emph{32nd USENIX Security Symposium (USENIX Security 23)}, pages
  5109--5126, Anaheim, CA, 2023. USENIX Association.
\newblock URL
  \url{https://www.usenix.org/conference/usenixsecurity23/presentation/eisenhofer}.

\bibitem[Eisenstein et~al.(2024)Eisenstein, Nagpal, Agarwal, Beirami, D'Amour,
  Dvijotham, Fisch, Heller, Pfohl, Ramachandran, Shaw, and
  Berant]{eisenstein2024}
Jacob Eisenstein, Chirag Nagpal, Alekh Agarwal, Ahmad Beirami, Alex D'Amour,
  DJ~Dvijotham, Adam Fisch, Katherine Heller, Stephen Pfohl, Deepak
  Ramachandran, Peter Shaw, and Jonathan Berant.
\newblock Helping or herding? {R}eward model ensembles mitigate but do not
  eliminate reward hacking.
\newblock In \emph{Conference on Language Modeling (COLM)}, 2024.
\newblock URL \url{https://openreview.net/forum?id=5u1GpUkKtG}.

\bibitem[Gao et~al.(2023)Gao, Schulman, and Hilton]{rewardoveropt}
Leo Gao, John Schulman, and Jacob Hilton.
\newblock Scaling laws for reward model overoptimization.
\newblock In \emph{Proceedings of the 40th International Conference on Machine
  Learning}, volume 202 of \emph{Proceedings of Machine Learning Research},
  pages 10835--10866. PMLR, 2023.

\bibitem[Gardner et~al.(2020)Gardner, Artzi, Basmov, Berant, Bogin, Chen,
  Dasigi, Dua, Elazar, Gottumukkala, Gupta, Hajishirzi, Ilharco, Khashabi, Lin,
  Liu, Liu, Mulcaire, Ning, Singh, Smith, Subramanian, Tsarfaty, Wallace,
  Zhang, and Zhou]{contrastsets}
Matt Gardner, Yoav Artzi, Victoria Basmov, Jonathan Berant, Ben Bogin, Sihao
  Chen, Pradeep Dasigi, Dheeru Dua, Yanai Elazar, Ananth Gottumukkala, Nitish
  Gupta, Hannaneh Hajishirzi, Gabriel Ilharco, Daniel Khashabi, Kevin Lin,
  Jiangming Liu, Nelson~F. Liu, Phoebe Mulcaire, Qiang Ning, Sameer Singh,
  Noah~A. Smith, Sanjay Subramanian, Reut Tsarfaty, Eric Wallace, Ally Zhang,
  and Ben Zhou.
\newblock Evaluating models' local decision boundaries via contrast sets.
\newblock In \emph{Findings of the Association for Computational Linguistics:
  EMNLP 2020}, pages 1307--1323. Association for Computational Linguistics,
  2020.
\newblock \doi{10.18653/v1/2020.findings-emnlp.117}.

\bibitem[{ggml-org}(2026)]{llamacpp}
{ggml-org}.
\newblock llama.cpp: {LLM} inference in {C}/{C++}.
\newblock \url{https://github.com/ggml-org/llama.cpp}, 2026.
\newblock Accessed 29 October 2026.

\bibitem[Han et~al.(2025)Han, Yang, and Kim]{exampleA}
Yewon Han, Seoyun Yang, and Taesup Kim.
\newblock Contrastive residual energy test-time adaptation.
\newblock \emph{arXiv preprint arXiv:2505.19607}, 2025.
\newblock URL \url{https://arxiv.org/abs/2505.19607}.
\newblock Licensed CC BY 4.0.

\bibitem[Holm(1979)]{holm}
Sture Holm.
\newblock A simple sequentially rejective multiple test procedure.
\newblock \emph{Scandinavian Journal of Statistics}, 6\penalty0 (2):\penalty0
  65--70, 1979.
\newblock URL \url{https://www.jstor.org/stable/4615733}.

\bibitem[Jia and Liang(2017)]{jialiang}
Robin Jia and Percy Liang.
\newblock Adversarial examples for evaluating reading comprehension systems.
\newblock In \emph{Proceedings of the 2017 Conference on Empirical Methods in
  Natural Language Processing}, pages 2021--2031. Association for Computational
  Linguistics, 2017.
\newblock \doi{10.18653/v1/D17-1215}.

\bibitem[Jiang et~al.(2022)Jiang, Xu, and Stevens]{arxivedits}
Chao Jiang, Wei Xu, and Samuel Stevens.
\newblock {arXivEdits}: Understanding the human revision process in scientific
  writing.
\newblock In \emph{Proceedings of the 2022 Conference on Empirical Methods in
  Natural Language Processing}, pages 9420--9435. Association for Computational
  Linguistics, 2022.
\newblock \doi{10.18653/v1/2022.emnlp-main.641}.

\bibitem[Jin et~al.(2020)Jin, Jin, Zhou, and Szolovits]{textfooler}
Di~Jin, Zhijing Jin, Joey~Tianyi Zhou, and Peter Szolovits.
\newblock Is {BERT} really robust? {A} strong baseline for natural language
  attack on text classification and entailment.
\newblock \emph{Proceedings of the AAAI Conference on Artificial Intelligence},
  34\penalty0 (05):\penalty0 8018--8025, 2020.
\newblock \doi{10.1609/aaai.v34i05.6311}.

\bibitem[Kang et~al.(2018)Kang, Ammar, Dalvi, van Zuylen, Kohlmeier, Hovy, and
  Schwartz]{peerread}
Dongyeop Kang, Waleed Ammar, Bhavana Dalvi, Madeleine van Zuylen, Sebastian
  Kohlmeier, Eduard Hovy, and Roy Schwartz.
\newblock A dataset of peer reviews ({PeerRead}): Collection, insights and
  {NLP} applications.
\newblock In \emph{Proceedings of the 2018 Conference of the North American
  Chapter of the Association for Computational Linguistics: Human Language
  Technologies, Volume 1 (Long Papers)}, pages 1647--1661. Association for
  Computational Linguistics, 2018.
\newblock \doi{10.18653/v1/n18-1149}.

\bibitem[Kaushik et~al.(2020)Kaushik, Hovy, and Lipton]{cad}
Divyansh Kaushik, Eduard Hovy, and Zachary~C. Lipton.
\newblock Learning the difference that makes a difference with
  counterfactually-augmented data.
\newblock In \emph{International Conference on Learning Representations}, 2020.
\newblock URL \url{https://openreview.net/forum?id=Sklgs0NFvr}.

\bibitem[Konan et~al.(2026{\natexlab{a}})Konan, Liu, and Liu]{cardarxiv}
Sachin Konan, Jonathan Liu, and Zhuang Liu.
\newblock {PaperScorer-3B-Text-arXiv}: Model card.
\newblock \url{https://huggingface.co/skonan/PaperScorer-3B-Text-arXiv},
  2026{\natexlab{a}}.
\newblock Accessed 2 November 2026.

\bibitem[Konan et~al.(2026{\natexlab{b}})Konan, Liu, and Liu]{cardiclr}
Sachin Konan, Jonathan Liu, and Zhuang Liu.
\newblock {PaperScorer-3B-Text-OpenReview-ICLR}: Model card.
\newblock
  \url{https://huggingface.co/skonan/PaperScorer-3B-Text-OpenReview-ICLR},
  2026{\natexlab{b}}.
\newblock Accessed 2 November 2026.

\bibitem[Konan et~al.(2026{\natexlab{c}})Konan, Liu, and Liu]{paperscorer}
Sachin Konan, Jonathan Liu, and Zhuang Liu.
\newblock {PaperScorer: How Predictable Is Paper Acceptance?}
\newblock \emph{arXiv preprint arXiv:2610.31791}, 2026{\natexlab{c}}.
\newblock URL \url{https://arxiv.org/abs/2610.31791}.

\bibitem[Koo et~al.(2024)Koo, Lee, Raheja, Park, Kim, and Kang]{koobias}
Ryan Koo, Minhwa Lee, Vipul Raheja, Jong~Inn Park, Zae~Myung Kim, and Dongyeop
  Kang.
\newblock Benchmarking cognitive biases in large language models as evaluators.
\newblock In \emph{Findings of the Association for Computational Linguistics:
  ACL 2024}, pages 517--545. Association for Computational Linguistics, 2024.
\newblock \doi{10.18653/v1/2024.findings-acl.29}.

\bibitem[Kuznetsov et~al.(2024)Kuznetsov, Afzal, Dercksen, Dycke, Goldberg,
  Hope, Hovy, Kummerfeld, Lauscher, Leyton-Brown, Lu, Mausam, Mieskes,
  N{\'e}v{\'e}ol, Pruthi, Qu, Schwartz, Smith, Solorio, Wang, Zhu, Rogers,
  Shah, and Gurevych]{nlp4review}
Ilia Kuznetsov, Osama~Mohammed Afzal, Koen Dercksen, Nils Dycke, Alexander
  Goldberg, Tom Hope, Dirk Hovy, Jonathan~K. Kummerfeld, Anne Lauscher, Kevin
  Leyton-Brown, Sheng Lu, Mausam, Margot Mieskes, Aur{\'e}lie N{\'e}v{\'e}ol,
  Danish Pruthi, Lizhen Qu, Roy Schwartz, Noah~A. Smith, Thamar Solorio,
  Jingyan Wang, Xiaodan Zhu, Anna Rogers, Nihar~B. Shah, and Iryna Gurevych.
\newblock What can natural language processing do for peer review?
\newblock \emph{arXiv preprint arXiv:2405.06563}, 2024.
\newblock URL \url{https://arxiv.org/abs/2405.06563}.

\bibitem[Kwon et~al.(2023)Kwon, Li, Zhuang, Sheng, Zheng, Yu, Gonzalez, Zhang,
  and Stoica]{vllm}
Woosuk Kwon, Zhuohan Li, Siyuan Zhuang, Ying Sheng, Lianmin Zheng, Cody~Hao Yu,
  Joseph Gonzalez, Hao Zhang, and Ion Stoica.
\newblock Efficient memory management for large language model serving with
  {PagedAttention}.
\newblock In \emph{Proceedings of the 29th Symposium on Operating Systems
  Principles}, pages 611--626, 2023.
\newblock \doi{10.1145/3600006.3613165}.

\bibitem[Li et~al.(2026{\natexlab{a}})Li, Jung, Zhao, Liu, Wang, Xu, Guo, Chen,
  Fei, Ling, and Bai]{autosupervision}
Haobo Li, Eunseo Jung, Wenxiao Zhao, Feng Liu, Jiong Wang, Kaiyi Xu, Zijie Guo,
  Zixin Chen, Ben Fei, Fenghua Ling, and Lei Bai.
\newblock {AutoSupervision: Closing the Feedback Loop in Scientific Workflows
  with Grounded Revision Verification}.
\newblock \emph{arXiv preprint arXiv:2607.27845}, 2026{\natexlab{a}}.
\newblock URL \url{https://arxiv.org/abs/2607.27845}.

\bibitem[Li et~al.(2026{\natexlab{b}})Li, Zhang, Davies, Qiu, and
  Gal]{abstractattack}
Lin Li, Qi~Zhang, Xander Davies, Jianing Qiu, and Yarin Gal.
\newblock {Gaming AI-Assisted Peer Reviews Poses New Risks to the Scientific
  Community}.
\newblock \emph{arXiv preprint arXiv:2606.10159}, 2026{\natexlab{b}}.
\newblock URL \url{https://arxiv.org/abs/2606.10159}.

\bibitem[Li et~al.(2026{\natexlab{c}})Li, Wang, Li, Zeng, Li, Shi, Zhou, and
  Zhou]{rhetoric}
Ming Li, Chenguang Wang, Xirui Li, Xinyue Zeng, Dianqi Li, Peng Shi, Dawei
  Zhou, and Tianyi Zhou.
\newblock {How Can Rhetoric Reward-Hack AI Reviewers? Dissecting Rhetorical
  Sensitivity in AI-Based Peer Review}.
\newblock \emph{arXiv preprint arXiv:2608.08975}, 2026{\natexlab{c}}.
\newblock URL \url{https://arxiv.org/abs/2608.08975}.

\bibitem[Liang et~al.(2024)Liang, Zhang, Cao, Wang, Ding, Yang, Vodrahalli, He,
  Smith, Yin, McFarland, and Zou]{liangfeedback}
Weixin Liang, Yuhui Zhang, Hancheng Cao, Binglu Wang, Daisy~Yi Ding, Xinyu
  Yang, Kailas Vodrahalli, Siyu He, Daniel~Scott Smith, Yian Yin, Daniel~A.
  McFarland, and James Zou.
\newblock Can large language models provide useful feedback on research papers?
  {A} large-scale empirical analysis.
\newblock \emph{NEJM AI}, 1\penalty0 (8), 2024.
\newblock \doi{10.1056/aioa2400196}.

\bibitem[Liu et~al.(2023)Liu, Iter, Xu, Wang, Xu, and Zhu]{geval}
Yang Liu, Dan Iter, Yichong Xu, Shuohang Wang, Ruochen Xu, and Chenguang Zhu.
\newblock {G-Eval}: {NLG} evaluation using {GPT-4} with better human alignment.
\newblock In \emph{Proceedings of the 2023 Conference on Empirical Methods in
  Natural Language Processing}, pages 2511--2522. Association for Computational
  Linguistics, 2023.
\newblock \doi{10.18653/v1/2023.emnlp-main.153}.

\bibitem[Manheim and Garrabrant(2018)]{goodhart}
David Manheim and Scott Garrabrant.
\newblock Categorizing variants of {G}oodhart's law.
\newblock \emph{arXiv preprint arXiv:1803.04585}, 2018.
\newblock URL \url{https://arxiv.org/abs/1803.04585}.

\bibitem[Marx et~al.(2020)Marx, Calmon, and Ustun]{marx2020}
Charles Marx, Flavio Calmon, and Berk Ustun.
\newblock Predictive multiplicity in classification.
\newblock In \emph{Proceedings of the 37th International Conference on Machine
  Learning}, volume 119 of \emph{Proceedings of Machine Learning Research},
  pages 6765--6774, 2020.
\newblock URL \url{https://proceedings.mlr.press/v119/marx20a.html}.

\bibitem[McCoy et~al.(2020)McCoy, Min, and Linzen]{mccoy2020}
R.~Thomas McCoy, Junghyun Min, and Tal Linzen.
\newblock {BERT}s of a feather do not generalize together: Large variability in
  generalization across models with similar test set performance.
\newblock In \emph{Proceedings of the Third BlackboxNLP Workshop on Analyzing
  and Interpreting Neural Networks for NLP}, pages 217--227. Association for
  Computational Linguistics, 2020.
\newblock \doi{10.18653/v1/2020.blackboxnlp-1.21}.

\bibitem[Morris et~al.(2020)Morris, Lifland, Yoo, Grigsby, Jin, and
  Qi]{textattack}
John Morris, Eli Lifland, Jin~Yong Yoo, Jake Grigsby, Di~Jin, and Yanjun Qi.
\newblock {TextAttack}: A framework for adversarial attacks, data augmentation,
  and adversarial training in {NLP}.
\newblock In \emph{Proceedings of the 2020 Conference on Empirical Methods in
  Natural Language Processing: System Demonstrations}, pages 119--126.
  Association for Computational Linguistics, 2020.
\newblock \doi{10.18653/v1/2020.emnlp-demos.16}.

\bibitem[Panickssery et~al.(2024)Panickssery, Bowman, and Feng]{selfpref}
Arjun Panickssery, Samuel~R. Bowman, and Shi Feng.
\newblock {LLM} evaluators recognize and favor their own generations.
\newblock In \emph{Advances in Neural Information Processing Systems 37}, 2024.
\newblock \doi{10.52202/079017-2197}.
\newblock URL
  \url{https://proceedings.neurips.cc/paper_files/paper/2024/hash/7f1f0218e45f5414c79c0679633e47bc-Abstract-Conference.html}.

\bibitem[Papernot et~al.(2017)Papernot, McDaniel, Goodfellow, Jha, Celik, and
  Swami]{blackbox}
Nicolas Papernot, Patrick McDaniel, Ian Goodfellow, Somesh Jha, Z.~Berkay
  Celik, and Ananthram Swami.
\newblock Practical black-box attacks against machine learning.
\newblock In \emph{Proceedings of the 2017 ACM on Asia Conference on Computer
  and Communications Security}, pages 506--519. ACM, 2017.
\newblock \doi{10.1145/3052973.3053009}.

\bibitem[Paszke et~al.(2019)Paszke, Gross, Massa, Lerer, Bradbury, Chanan,
  et~al.]{pytorch}
Adam Paszke, Sam Gross, Francisco Massa, Adam Lerer, James Bradbury, Gregory
  Chanan, et~al.
\newblock {PyTorch}: An imperative style, high-performance deep learning
  library.
\newblock In \emph{Advances in Neural Information Processing Systems 32}, 2019.
\newblock URL
  \url{https://papers.nips.cc/paper_files/paper/2019/hash/bdbca288fee7f92f2bfa9f7012727740-Abstract.html}.

\bibitem[Pedregosa et~al.(2011)Pedregosa, Varoquaux, Gramfort, Michel, Thirion,
  Grisel, et~al.]{sklearn}
Fabian Pedregosa, Ga{\"e}l Varoquaux, Alexandre Gramfort, Vincent Michel,
  Bertrand Thirion, Olivier Grisel, et~al.
\newblock Scikit-learn: Machine learning in {P}ython.
\newblock \emph{Journal of Machine Learning Research}, 12\penalty0
  (85):\penalty0 2825--2830, 2011.
\newblock URL \url{https://jmlr.org/papers/v12/pedregosa11a.html}.

\bibitem[Powers et~al.(2002)Powers, Burstein, Chodorow, Fowles, and
  Kukich]{powers2002}
Donald~E. Powers, Jill~C. Burstein, Martin Chodorow, Mary~E. Fowles, and Karen
  Kukich.
\newblock Stumping e-rater: Challenging the validity of automated essay
  scoring.
\newblock \emph{Computers in Human Behavior}, 18\penalty0 (2):\penalty0
  103--134, 2002.
\newblock \doi{10.1016/S0747-5632(01)00052-8}.

\bibitem[Qiu et~al.(2025)Qiu, Chen, and Li]{rhetoricalstyle}
Jingyi Qiu, Hong Chen, and Zongyi Li.
\newblock {Counterfactual LLM-based Framework for Measuring Rhetorical Style}.
\newblock \emph{arXiv preprint arXiv:2512.19908}, 2025.
\newblock URL \url{https://arxiv.org/abs/2512.19908}.

\bibitem[{Qwen} et~al.(2025)]{qwen25}
{Qwen} et~al.
\newblock {Qwen2.5} technical report.
\newblock \emph{arXiv preprint arXiv:2412.15115}, 2025.
\newblock URL \url{https://arxiv.org/abs/2412.15115v2}.

\bibitem[Ribeiro et~al.(2016)Ribeiro, Singh, and Guestrin]{lime}
Marco~Tulio Ribeiro, Sameer Singh, and Carlos Guestrin.
\newblock ``why should {I} trust you?'': Explaining the predictions of any
  classifier.
\newblock In \emph{Proceedings of the 22nd ACM SIGKDD International Conference
  on Knowledge Discovery and Data Mining}, pages 1135--1144. ACM, 2016.
\newblock \doi{10.1145/2939672.2939778}.

\bibitem[Ribeiro et~al.(2020)Ribeiro, Wu, Guestrin, and Singh]{checklist}
Marco~Tulio Ribeiro, Tongshuang Wu, Carlos Guestrin, and Sameer Singh.
\newblock Beyond accuracy: Behavioral testing of {NLP} models with {CheckList}.
\newblock In \emph{Proceedings of the 58th Annual Meeting of the Association
  for Computational Linguistics}, pages 4902--4912. Association for
  Computational Linguistics, 2020.
\newblock \doi{10.18653/v1/2020.acl-main.442}.

\bibitem[Skalse et~al.(2022)Skalse, Howe, Krasheninnikov, and
  Krueger]{rewardhacking}
Joar Skalse, Nikolaus H.~R. Howe, Dmitrii Krasheninnikov, and David Krueger.
\newblock Defining and characterizing reward gaming.
\newblock In \emph{Advances in Neural Information Processing Systems 35}, pages
  9460--9471, 2022.
\newblock \doi{10.52202/068431-0687}.
\newblock URL
  \url{https://proceedings.neurips.cc/paper_files/paper/2022/hash/3d719fee332caa23d5038b8a90e81796-Abstract-Conference.html}.

\bibitem[Spearman(1904)]{spearman1904}
C.~Spearman.
\newblock The proof and measurement of association between two things.
\newblock \emph{The American Journal of Psychology}, 15\penalty0 (1):\penalty0
  72--101, 1904.
\newblock \doi{10.2307/1412159}.
\newblock URL \url{https://www.jstor.org/stable/1412159}.

\bibitem[Stiennon et~al.(2020)Stiennon, Ouyang, Wu, Ziegler, Lowe, Voss,
  Radford, Amodei, and Christiano]{summarizehf}
Nisan Stiennon, Long Ouyang, Jeff Wu, Daniel~M. Ziegler, Ryan Lowe, Chelsea
  Voss, Alec Radford, Dario Amodei, and Paul Christiano.
\newblock Learning to summarize with human feedback.
\newblock In \emph{Advances in Neural Information Processing Systems 33}, pages
  3008--3021, 2020.
\newblock URL
  \url{https://proceedings.neurips.cc/paper/2020/hash/1f89885d556929e98d3ef9b86448f951-Abstract.html}.

\bibitem[Stureborg et~al.(2024)Stureborg, Alikaniotis, and
  Suhara]{inconsistent}
Rickard Stureborg, Dimitris Alikaniotis, and Yoshi Suhara.
\newblock Large language models are inconsistent and biased evaluators.
\newblock \emph{arXiv preprint arXiv:2405.01724}, 2024.
\newblock URL \url{https://arxiv.org/abs/2405.01724}.

\bibitem[Tram{\`e}r et~al.(2016)Tram{\`e}r, Zhang, Juels, Reiter, and
  Ristenpart]{tramer2016}
Florian Tram{\`e}r, Fan Zhang, Ari Juels, Michael~K. Reiter, and Thomas
  Ristenpart.
\newblock Stealing machine learning models via prediction {APIs}.
\newblock In \emph{25th USENIX Security Symposium (USENIX Security 16)}, pages
  601--618. USENIX Association, 2016.
\newblock URL
  \url{https://www.usenix.org/conference/usenixsecurity16/technical-sessions/presentation/tramer}.

\bibitem[Tram{\`e}r et~al.(2017)Tram{\`e}r, Papernot, Goodfellow, Boneh, and
  McDaniel]{transferspace}
Florian Tram{\`e}r, Nicolas Papernot, Ian Goodfellow, Dan Boneh, and Patrick
  McDaniel.
\newblock The space of transferable adversarial examples.
\newblock \emph{arXiv preprint arXiv:1704.03453}, 2017.
\newblock URL \url{https://arxiv.org/abs/1704.03453}.

\bibitem[{vLLM Project}(2026)]{vllmdocs}
{vLLM Project}.
\newblock vllm serve: command-line reference, option \texttt{--dtype}.
\newblock \url{https://docs.vllm.ai/en/stable/cli/serve/}, 2026.
\newblock Accessed 1 November 2026.

\bibitem[Wallace et~al.(2019)Wallace, Feng, Kandpal, Gardner, and
  Singh]{triggers}
Eric Wallace, Shi Feng, Nikhil Kandpal, Matt Gardner, and Sameer Singh.
\newblock Universal adversarial triggers for attacking and analyzing {NLP}.
\newblock In \emph{Proceedings of the 2019 Conference on Empirical Methods in
  Natural Language Processing and the 9th International Joint Conference on
  Natural Language Processing (EMNLP-IJCNLP)}, pages 2153--2162. Association
  for Computational Linguistics, 2019.
\newblock \doi{10.18653/v1/D19-1221}.

\bibitem[Wang et~al.(2026)Wang, Li, Fan, Chen, Chen, Zhou, and
  Zhou]{robustreview}
Chenguang Wang, Ming Li, Chengrui Fan, Jianpeng Chen, Han Chen, Tianyi Zhou,
  and Dawei Zhou.
\newblock {A Missing Piece for Trustworthy AI Reviewers: From Benchmarking
  Rhetorical Robustness to SciCore Review}.
\newblock \emph{arXiv preprint arXiv:2609.39027}, 2026.
\newblock URL \url{https://arxiv.org/abs/2609.39027}.

\bibitem[Wang et~al.(2024)Wang, Li, Chen, Cai, Zhu, Lin, Cao, Kong, Liu, Liu,
  and Sui]{unfairevaluators}
Peiyi Wang, Lei Li, Liang Chen, Zefan Cai, Dawei Zhu, Binghuai Lin, Yunbo Cao,
  Lingpeng Kong, Qi~Liu, Tianyu Liu, and Zhifang Sui.
\newblock Large language models are not fair evaluators.
\newblock In \emph{Proceedings of the 62nd Annual Meeting of the Association
  for Computational Linguistics (Volume 1: Long Papers)}, pages 9440--9450.
  Association for Computational Linguistics, 2024.
\newblock \doi{10.18653/v1/2024.acl-long.511}.

\bibitem[{WestlakeNLP}(2025)]{deepreviewer}
{WestlakeNLP}.
\newblock {DeepReviewer-14B: Model Card}, 2025.
\newblock URL \url{https://huggingface.co/WestlakeNLP/DeepReviewer-14B}.
\newblock Accessed 29 October 2026.

\bibitem[Wolf et~al.(2020)Wolf, Debut, Sanh, et~al.]{transformers}
Thomas Wolf, Lysandre Debut, Victor Sanh, et~al.
\newblock Transformers: State-of-the-art natural language processing.
\newblock In \emph{Proceedings of the 2020 Conference on Empirical Methods in
  Natural Language Processing: System Demonstrations}, pages 38--45.
  Association for Computational Linguistics, 2020.
\newblock \doi{10.18653/v1/2020.emnlp-demos.6}.

\bibitem[Wu et~al.(2026)Wu, Tang, Ma, Liu, Zheng, and Hao]{exampleB}
Zihao Wu, Hongyao Tang, Yi~Ma, Jiashun Liu, Yan Zheng, and Jianye Hao.
\newblock {The Rank and Gradient Lost in Non-stationarity: Sample Weight Decay
  for Mitigating Plasticity Loss in Reinforcement Learning}.
\newblock In \emph{International Conference on Learning Representations
  (ICLR)}, 2026.
\newblock URL
  \url{https://proceedings.iclr.cc/paper_files/paper/2026/hash/2886558a03d11f95d4020a460fe4a390-Abstract-Conference.html}.
\newblock Example text is available as arXiv:2604.01913 under CC BY 4.0.

\bibitem[Xiao et~al.(2024)Xiao, Liu, Zhang, Muennighoff, Lian, and Nie]{cpack}
Shitao Xiao, Zheng Liu, Peitian Zhang, Niklas Muennighoff, Defu Lian, and
  Jian-Yun Nie.
\newblock {C-Pack}: Packed resources for general {Chinese} embeddings.
\newblock In \emph{Proceedings of the 47th International ACM SIGIR Conference
  on Research and Development in Information Retrieval}, pages 641--649.
  Association for Computing Machinery, 2024.
\newblock \doi{10.1145/3626772.3657878}.

\bibitem[Yang et~al.(2026{\natexlab{a}})Yang, Sha, Li, Yu, Sun, Zhao, Fang,
  Guo, Wu, Hu, Luo, Liu, and Wang]{repackaging}
Xu~Yang, Zhizhou Sha, Junbo Li, Jian Yu, Yifan Sun, Matthew Zhao, Jinrui Fang,
  Xinyue Guo, Yining Wu, Xu~Hu, Yifu Luo, Qiang Liu, and Zhangyang Wang.
\newblock {No Hidden Prompts Needed! You Can Game AI Peer Review with
  Presentation-Only Revisions}.
\newblock \emph{arXiv preprint arXiv:2606.13044}, 2026{\natexlab{a}}.
\newblock URL \url{https://arxiv.org/abs/2606.13044}.

\bibitem[Yang et~al.(2026{\natexlab{b}})Yang, Li, Wang, Dobbie, and
  Liu]{gainagreement}
Zongrui Yang, Haoyuan Li, Zhongsheng Wang, Gillian Dobbie, and Jiamou Liu.
\newblock {Can We Believe AI Peer Review?}
\newblock \emph{arXiv preprint arXiv:2610.19339}, 2026{\natexlab{b}}.
\newblock URL \url{https://arxiv.org/abs/2610.19339}.

\bibitem[Yuan et~al.(2022)Yuan, Liu, and Neubig]{yuanreview}
Weizhe Yuan, Pengfei Liu, and Graham Neubig.
\newblock Can we automate scientific reviewing?
\newblock \emph{Journal of Artificial Intelligence Research}, 75:\penalty0
  171--212, 2022.
\newblock \doi{10.1613/jair.1.12862}.

\bibitem[Zhao et~al.(2026)Zhao, Khan, Xu, Tan, and Chen]{paperguard}
Xinyu Zhao, Rana Muhammad~Shahroz Khan, Zhen Xu, Zhen Tan, and Tianlong Chen.
\newblock {Does AI Reviewer See the Full Picture? Attacking and Defending
  Multimodal Peer Review}.
\newblock In \emph{Proceedings of the 43rd International Conference on Machine
  Learning}, volume 306 of \emph{Proceedings of Machine Learning Research},
  pages 162054--162074, 2026.
\newblock URL \url{https://proceedings.mlr.press/v306/zhao26an.html}.

\bibitem[Zheng et~al.(2023)Zheng, Chiang, Sheng, Zhuang, Wu, Zhuang, Lin, Li,
  Li, Xing, Zhang, Gonzalez, and Stoica]{mtbench}
Lianmin Zheng, Wei-Lin Chiang, Ying Sheng, Siyuan Zhuang, Zhanghao Wu, Yonghao
  Zhuang, Zi~Lin, Zhuohan Li, Dacheng Li, Eric~P. Xing, Hao Zhang, Joseph~E.
  Gonzalez, and Ion Stoica.
\newblock Judging {LLM}-as-a-judge with {MT}-bench and chatbot arena.
\newblock In \emph{Advances in Neural Information Processing Systems 36,
  Datasets and Benchmarks Track}, pages 46595--46623, 2023.
\newblock \doi{10.52202/075280-2020}.
\newblock URL
  \url{https://proceedings.neurips.cc/paper_files/paper/2023/hash/91f18a1287b398d378ef22505bf41832-Abstract-Datasets_and_Benchmarks.html}.

\bibitem[Zhou et~al.(2026)Zhou, Zhang, Zhi, and Sun]{injection}
Qin Zhou, Zhexin Zhang, Li~Zhi, and Limin Sun.
\newblock ``{G}ive a positive review only'': An early investigation into
  in-paper prompt injection attacks and defenses for {AI} reviewers.
\newblock In \emph{Findings of the Association for Computational Linguistics:
  EMNLP 2026}, pages 982--997, Budapest, Hungary, 2026. Association for
  Computational Linguistics.
\newblock \doi{10.18653/2026.findings-emnlp.49}.
\newblock URL \url{https://aclanthology.org/2026.findings-emnlp.49/}.

\bibitem[Zhou et~al.(2024)Zhou, Chen, and Yu]{reliablereviewer}
Ruiyang Zhou, Lu~Chen, and Kai Yu.
\newblock Is {LLM} a reliable reviewer? {A} comprehensive evaluation of {LLM}
  on automatic paper reviewing tasks.
\newblock In \emph{Proceedings of the 2024 Joint International Conference on
  Computational Linguistics, Language Resources and Evaluation (LREC-COLING
  2024)}, pages 9340--9351. European Language Resources Association (ELRA) and
  ICCL, 2024.
\newblock \doi{10.63317/48d359hjdvog}.

\bibitem[Ziegler et~al.(2019)Ziegler, Stiennon, Wu, Brown, Radford, Amodei,
  Christiano, and Irving]{ziegler}
Daniel~M. Ziegler, Nisan Stiennon, Jeffrey Wu, Tom~B. Brown, Alec Radford,
  Dario Amodei, Paul Christiano, and Geoffrey Irving.
\newblock Fine-tuning language models from human preferences.
\newblock \emph{arXiv preprint arXiv:1909.08593}, 2019.
\newblock URL \url{https://arxiv.org/abs/1909.08593}.

\end{thebibliography}
\clearpage
\section*{Supplementary Material}
\phantomsection
\addcontentsline{toc}{section}{Supplementary Material}
\setcounter{section}{0}
\setcounter{table}{0}
\setcounter{figure}{0}
\setcounter{equation}{0}
\renewcommand{\thesection}{S\arabic{section}}
\renewcommand{\thetable}{S\arabic{table}}
\renewcommand{\thefigure}{S\arabic{figure}}
\renewcommand{\theequation}{S\arabic{equation}}
\renewcommand{\theHsection}{S\arabic{section}}
\renewcommand{\theHtable}{S\arabic{table}}
\renewcommand{\theHfigure}{S\arabic{figure}}
\renewcommand{\theHequation}{S\arabic{equation}}
\newcommand{\appendixsection}[1]{\section{#1}}
This appendix supplies supplementary methods and results. Section, table, figure and equation numbers prefixed with S refer to this appendix; unprefixed numbers refer to the article. Here $S$ is the sparse classifier; $T_1$ and $T_2$ are the two PaperScorer acceptance models trained on International Conference on Learning Representations (ICLR) and arXiv papers, respectively. AUC denotes area under the receiver operating characteristic curve; SD denotes standard deviation; CI denotes confidence interval. CPU and GPU denote central and graphics processing units, respectively.
\begin{figure}[H]
\centering
\includegraphics[width=\linewidth]{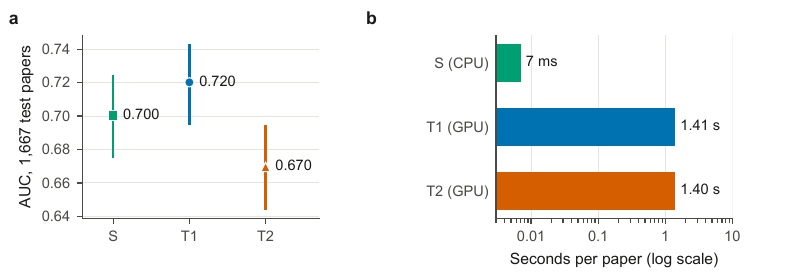}
\caption{Prediction and inference-cost context. AUC denotes area under the receiver operating characteristic curve; CI denotes confidence interval; CPU and GPU denote central and graphics processing units. (a)~AUC on the common 1,667-paper original-text benchmark, with nominal 95\% CIs. (b)~Median inference time from extracted text, on a CPU for \Surr{} and one RTX 4090 GPU for \Tone{} and \Ttwo. The CPU median uses the first 200 benchmark papers; target medians use all 5,321 frozen inputs. Different processors and timing cohorts make this a practical cost comparison, not an isolated architectural speedup measurement.}
\label{fig:context}
\end{figure}
\appendixsection{Evaluator configuration and input handling}
\label{app:surrogate}
The lexical vocabulary contains at most 300,000 features, using word unigrams and bigrams with minimum document frequency three and sublinear term frequency. Vocabulary fitting uses training papers only. Structural features are log character count, figure-marker count, table count, displayed-equation count, heading count, citation-pattern count, title character count, and abstract character count. They are standardized using training-set means and variances. A GitHub-link indicator present in the initial feature inventory is removed before fitting the selected surrogate.

The structural block is multiplied by a weight chosen from $\{0.05,0.1,0.3\}$. Logistic regression selects its inverse regularization strength from $\{0.01,0.03,0.1,0.3,1,3,10\}$ and runs at most 3,000 solver iterations. Validation AUC selects weight 0.3 and inverse regularization strength 1. The resulting validation AUC is 0.691. Training uses acceptance decisions rather than target-model scores. At inference, the abstract and structural features are recomputed for each edited document; lexical input is restricted to its first 30,000 characters. There is no online fitting during search.

The original-paper comparison contains 676 ICLR 2025 and 991 ICLR 2026 papers. Exact normalized-title matching finds 160 titles in \Ttwo's arXiv training split: 62 from 2025 and 98 from 2026. Thus, \Ttwo's full-benchmark AUC is context rather than a fully held-out estimate. The edit-study sampling separately excludes identified training families from both targets' training corpora. This larger context benchmark is distinct from the 860-paper family-filtered source pool used for the editing studies. In the same-GPU sibling comparison, target AUCs on ICLR 2026 alone are 0.696 for \Tone{} and 0.635 for \Ttwo. The latter's lower discrimination is another reason to avoid interpreting the two targets as interchangeable replications. A separate structure-only surrogate baseline reaches AUC 0.642; it is an ablation of our features, rather than an implementation of PaperScorer's feature inventory.

\subsection{Editing-study sampling and target execution}
Editing candidates come only from the 991 ICLR 2026 test papers. We compare normalized titles and title-plus-abstract term frequency--inverse document frequency (TF--IDF) vectors with both released training corpora. Exact title matches and cosine similarity of at least 0.80 are excluded; similarity from 0.60 to below 0.80 is set aside. This leaves 860 candidates before excluding the 30 pilot papers and selecting the disjoint study samples. The filter concerns identified paper families, not base-model pretraining exposure.

All paired experimental target scores use an NVIDIA RTX 4090 GPU. The forward pass uses bfloat16, a reduced-precision format, and the acceptance/rejection decision is read in 32-bit floating point. Each input receives one score per target; repeated edits are not independent papers.

\subsection{Target prompts and decision readout}
The shared target system message is:
\begin{quote}\small
You are an expert academic reviewer tasked with evaluating research papers.
\end{quote}
\par\noindent\begin{minipage}{\linewidth}
The ICLR user preamble is:
\begin{quote}\small
I am giving you a paper. I want to predict its acceptance outcome at ICLR.\\
- Your answer will either be: \verb|\boxed{Accept}| or \verb|\boxed{Reject}|\\
- Note: ICLR generally has a $\sim$30\% acceptance rate
\end{quote}
\end{minipage}\par
\par\noindent\begin{minipage}{\linewidth}
The arXiv user preamble is:
\begin{quote}\small
I am giving you a paper submitted to a top machine-learning venue. Predict its acceptance outcome.\\
- Your answer will either be: \verb|\boxed{Accept}| or \verb|\boxed{Reject}|\\
- Note: typical top-tier ML venues have $\sim$25--30\% acceptance rates
\end{quote}
\end{minipage}\par
The respective preamble is followed by the paper text. The model's chat template supplies an assistant-generation prefix, after which the readout prefix of the release's scoring rule is \verb|Outcome: \boxed{|. If $z_A$ and $z_R$ are the acceptance and rejection logits, the score of evaluator $a$ for paper text $x$ is
\begin{equation}
a(x)=\frac{\exp(z_A)}{\exp(z_A)+\exp(z_R)}.
\end{equation}
Both labels are single tokens in the released tokenizers. As an implementation choice of this study, the final hidden state and the two relevant output-head rows are converted to 32-bit floating point for this calculation, avoiding coarse probability ties from a bfloat16-only decision readout.

The full prompt limit is 24,480 tokens. The body allowance subtracts the tokenized preamble and chat-template overhead, with 32 tokens reserved for the decision prefix. Longer bodies are truncated before the final template is assembled. Prefill proceeds in chunks of 4,096 tokens using the key--value cache. Target scores in the main paired analyses all come from the RTX 4090. A post hoc RTX 5090 rescoring (Supplementary Section~\ref{app:reliability}) measures reliability only and is not mixed into experimental changes.

\subsection{Search algorithms and budget accounting}
\label{app:search_protocol}
Study 3 searches over admitted single-paragraph edits, allowing at most eight edits, one replacement per paragraph, and at most a 5\% change relative to the original total character count. Both policies score candidates only with \Surr. Adaptive search evaluates compatible one-edit extensions in seeded random order and greedily accepts the best improving extension. It stops when no extension improves the current score, no feasible extension remains, or the evaluation allowance is exhausted. Random search samples feasible composites with an edit count drawn uniformly from one to eight, scores them with \Surr, and retains the highest-scoring candidate. Thus, \emph{best-of-$Q$ random} means best according to \Surr, not best according to a target.

Both policies retain the unedited paper as an eligible choice, scored once outside the candidate allowance. We run two seeds per paper and record nested checkpoints at $Q\in\{1,8,64,512\}$. The policies share a maximum allowance, rather than necessarily consuming the same number of evaluations: adaptive search can stop early and carry its final choice forward. Distinct nonempty endpoints at 64 and 512 are checked as complete documents by the semantic checker. An invalid endpoint contributes zero gain for that policy, rather than being omitted. Endpoints at one and eight are constructed from admitted edits but are not checked as complete documents; these budgets are exploratory. The one-pass polish supplies a separate, non-search control.

\subsection{Primary score changes and transfer equation}
\label{app:estimands}
For an original paper $x_i$, evaluator $a$, and edit $e$, the score change is
\begin{equation}
\Delta a_{ie}=a(x_i^{(e)})-a(x_i).
\end{equation}
For the transfer estimand, let $n_3=128$ be the number of Study 3 papers, and let $A_{ir}(Q)$ and $R_{ir}(Q)$ be the endpoints chosen by adaptive and random search for paper $i$ and seed $r$. Let $\widetilde{\Delta}T$ equal the endpoint's target change if admitted, and zero if the endpoint is invalid or unedited. Then
\begin{equation}
G_T(Q)=\frac{1}{n_3}\sum_{i=1}^{n_3}\frac{1}{2}\sum_{r=0}^{1}
\frac{\widetilde{\Delta}T(A_{ir}(Q))-\widetilde{\Delta}T(R_{ir}(Q))}{\sigma_T}.
\label{eq:transfer}
\end{equation}
The scale $\sigma_a$ is the sample standard deviation (SD) of evaluator $a$ on all 328 frozen originals: 0.1530 for \Surr, 0.2266 for \Tone, and 0.2730 for \Ttwo{} in acceptance-score units. Standardization makes effects relative to each evaluator's between-paper variation; it does not convert a model score into scientific quality. Points are 100 times the raw acceptance score, not the release's calibrated estimate; 32-bit changes are converted with the 32-bit SDs over the 200 Study 2 originals, 0.2421 for \Tone{} and 0.2569 for \Ttwo; the 32-bit search contrast uses those over all 328 originals, 0.244 and 0.256.

\appendixsection{Editing and checking instructions}
\label{app:prompts}
This section reproduces the shared scientific editing constraints and the task instructions. Each call supplies the complete anonymized source paper, divided into numbered paragraphs. The rules below are instructions to the experimental writer, rather than guarantees about its output; validity rates in Section~\ref{sec:validity} quantify compliance.

\subsection{Shared writer rules}
\label{app:writer_rules}
\begin{quote}\small
Rules for every change (presentation only):
\begin{enumerate}
\item Keep every number, equation, citation, table, figure/table/section reference and heading exactly.
\item Claim strength is frozen: keep every hedge and intensifier exactly as written (for example: may, might, can, could, likely, suggests, aims to, attempts to, ensures, guarantees, significantly, substantially, highly, only, first, novel, unique, state-of-the-art, outperforms). Never turn an aim or attempt into an achievement, or the reverse. Never add or remove a statement that something is novel, first or unique.
\item Keep every factual statement and attribution; add or remove no results or evidence; do not move text between paragraphs.
\item You may change wording, sentence order within the paragraph, transitions, signposting and emphasis.
\item Keep the paper anonymous.
\end{enumerate}
\end{quote}
The evidence-removal and sham tasks explicitly authorize their declared deletions; the remaining constraints apply outside the deleted spans.

\begin{samepage}
\subsection{Atomic edits and deletions}
The bank-generation task is:
\begin{quote}\small
Write 24 atomic presentation edits. Each edit replaces exactly one paragraph [Pn] (all pids distinct; skip headings, tables, equations, captions and reference lists) with a version that reads more convincingly to reviewers; give its kind (clarity, framing, structure, emphasis or flow).
\end{quote}
\end{samepage}
\noindent\begin{minipage}{\linewidth}
For Study 2, the task continues:
\begin{quote}\small
Also define (a) an evidence removal: the paragraph ids that report exactly ONE experimental result supporting a claim stated in the abstract or introduction (quote that claim), removed by deleting those whole paragraphs and nothing else; (b) a sham: whole paragraphs from background or related work only, containing no results, no claim of this paper and no figure or table, whose total length is within 20\% of the removal's.
\end{quote}
\end{minipage}
The supplied output schema requires a paragraph number, replacement text, and edit type for each edit. Study 2 additionally requires the removal paragraph numbers and supporting claim, and the sham paragraph numbers. These are structured outputs; source manuscript text is not included in the planned public release.

\subsection{One-pass polishing}
\begin{quote}\small
Polish the paper's presentation in one pass, as a careful author would before submission. Return only the paragraphs you change, each with its pid and full new text; skip headings, tables, equations, captions and reference lists.
\end{quote}
The shared writer rules in Section~\ref{app:writer_rules} also apply. Each returned paragraph is checked separately, and only admitted replacements are composed into the scored polished version. This reduces content drift, while making the polishing intervention dependent on the checker as well as the writer.

\subsection{Semantic check}
The checker sees the original paper plus the original and changed spans of every proposed item. Its shared instruction is:
\begin{quote}\small
Check each ITEM. An edit is valid only if it is presentation-only: the same claims with the same strength and scope (no hedge or intensifier added, removed or changed; no aim turned into an achievement or the reverse; no novelty claim added or removed), the same numbers, citations and attributions, no added or removed evidence, nothing moved from other paragraphs.
\end{quote}
For deletions, the checker is additionally told that a removal must delete exactly one experimental result supporting the quoted claim and leave the paper coherent. A sham must remove only background or related work that is not evidence for the paper's claims and contains no figure or table. For composite endpoints, the introductory instruction states that individually approved edits must be judged together as a whole. The second provider receives the same scientific checking task without the first checker's verdicts or target scores.

\subsection{Deterministic invariants}
The checker compares multisets, rather than only exact whole-document strings. Numeric patterns include signs, decimal and grouped digits, and percentages. Equations are normalized for whitespace and braces around a single subscript or superscript token. Author--year citations and numeric citations are extracted separately. Cross-reference spelling is normalized, for example ``Fig.~1'' and ``Figure~1'', while preserving the referenced object and number. Markdown headings and Hypertext Markup Language (HTML) table blocks are also protected. For deletions, the declared whole paragraphs must occur in the original; only their protected tokens are permitted to disappear. These checks cannot detect all semantic changes, which is why semantic judgments and human sampling remain necessary.

\subsection{Worked examples}
\label{app:examples}
The two edits below were chosen by a rule fixed before any candidate was viewed. The pool contains admitted Study~2 single edits from papers that belong to the 198-paper fidelity cohort, are also on arXiv under a Creative Commons Attribution (CC~BY) license or CC0 public-domain dedication, and have a source paragraph of at most 120 words: 506 edits in 40 papers. Example~A is the edit whose absolute \Tone{} change is closest to the median absolute \Tone{} change over all admitted Study~2 edits, 0.056~SD. Example~B applies the same rule to edits on which \Surr{} and \Tone{} move in opposite directions, from a different paper. Example~A happens to show opposite directions as well. Both edits passed the invariant check and the blind semantic check. Score changes are in each evaluator's SD units. The quoted paragraphs are from the PaperScorer release of each ICLR 2026 submission; both also appear in the arXiv versions, which are licensed CC~BY~4.0, with different cross-reference numbers in Example~A. Two edits illustrate the measured object; they are not evidence about the mean relation, which Section~\ref{sec:fidelity} reports over all pairs.

\noindent\begin{minipage}{\linewidth}
\paragraph{Example A, a flow edit, from \citet{exampleA}}
\begin{quote}\small
\textbf{Original:} If we remove the residual assumption, the pairwise contrastive learning objective reduces to the form in subsection C.1. In this case, there are no bias terms parameterized by $\phi$, so the gradient magnitude depends solely on the target model's energies, making the method more prone to overfitting, as illustrated in Figure 3. A more detailed mathematical discussion of the residual assumption and the pairwise contrastive approach is provided in Appendix B and Appendix C.\par\smallskip\noindent\textbf{Edited:} Without the residual assumption, the pairwise contrastive learning objective reduces to the form in subsection C.1. In this case, there are no bias terms parameterized by $\phi$, so the gradient magnitude depends solely on the target model's energies, making the method more prone to overfitting, as illustrated in Figure 3. Appendix B and Appendix C provide a more detailed mathematical discussion of the residual assumption and the pairwise contrastive approach.\par\smallskip\noindent\textbf{Score changes:} \Surr{} $-0.023$, \Tone{} $+0.056$, \Ttwo{} $+0.045$ SD.
\end{quote}
\end{minipage}

\noindent\begin{minipage}{\linewidth}
In Example B, NTK denotes neural tangent kernel; GD and SGD denote gradient descent and stochastic gradient descent.

\paragraph{Example B, a structure edit, from \citet{exampleB}}
\begin{quote}\small
\textbf{Original:} Prior research (Du et al., 2019; Allen-Zhu et al., 2019) shows overparameterized networks achieve global convergence to zero training error via Gradient Descent (GD) or Stochastic Gradient Descent (SGD) if two conditions hold: (i) The initial NTK matrix $\mathbf{K}_0$ is well-conditioned: its eigenvalues are bounded (no extremely large/small values to distort optimization). (ii) The NTK matrix $\mathbf{K}$ remains stable during training, so its structural properties do not degrade and harm convergence.\par\smallskip\noindent\textbf{Edited:} Prior research (Du et al., 2019; Allen-Zhu et al., 2019) shows that overparameterized networks achieve global convergence to zero training error via Gradient Descent (GD) or Stochastic Gradient Descent (SGD) if two conditions hold. First, the initial NTK matrix $\mathbf{K}_0$ is well-conditioned, with eigenvalues bounded so that extremely large or small values do not distort optimization. Second, the NTK matrix $\mathbf{K}$ remains stable during training, so its structural properties do not degrade and harm convergence.\par\smallskip\noindent\textbf{Score changes:} \Surr{} $+0.011$, \Tone{} $-0.056$, \Ttwo{} $+0.010$ SD.
\end{quote}
\end{minipage}

\FloatBarrier
\appendixsection{Validity, sample accounting, and human judgments}
\label{app:validity}

\begin{figure}[htbp]
\centering
\includegraphics[width=\linewidth]{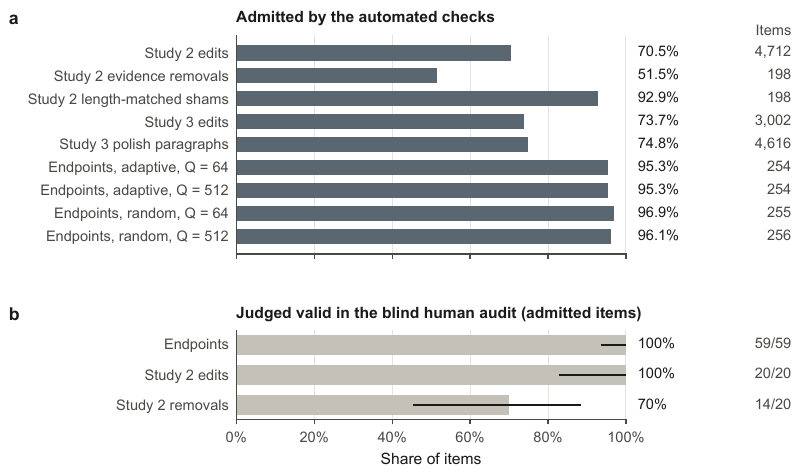}
\caption{Automated admission and blind human validity are different measurements. (a)~Share of proposed items admitted by the automated checks, with the number of checkable proposals at the right; composite endpoints receive a further whole-document semantic check. (b)~Share of admitted items judged valid in the later blind human audit, with the counts at the right and exact two-sided 95\% binomial confidence intervals (CIs); an unsure judgment counts as not valid. The audit samples 59 endpoints, 20 presentation edits, and 20 evidence removals. These are conditional samples of admitted items and cannot be compared directly with generation pass rates as if they shared a denominator. $Q$ is the maximum number of surrogate evaluations allowed to the search.}
\label{fig:validity}
\end{figure}

\begin{table}[htbp]
\centering
\caption{Validity of generated items. The last column requires both the deterministic check and the semantic checker to pass. Polish items are proposed paragraphs, not complete papers. Counts exclude absent, empty, or unchanged proposals that do not become checkable items.}
\label{tab:admission_rates}
\begin{tabular}{llrrr}
\toprule
Study & Item & Count & Invariants pass & Both pass \\
\midrule
2 & Presentation edit & 4,712 & 90.9\% & 70.5\% \\
2 & Evidence removal & 198 & 98.5\% & 51.5\% \\
2 & Deletion sham & 198 & 98.5\% & 92.9\% \\
3 & Presentation edit & 3,002 & 92.5\% & 73.7\% \\
3 & Polish paragraph & 4,616 & 92.9\% & 74.8\% \\
\bottomrule
\end{tabular}
\end{table}
There are 200 selected Study 2 originals and 128 Study 3 originals in the frozen set. Of the former, 199 have admitted single edits, and 198 enter the correlation analysis. The generated-item denominators in Table~\ref{tab:admission_rates} are checkable proposals, not the requested maximum of 24 times the number of papers. Evidence removals and shams each have 198 checkable proposals, of which 102 and 184 pass both automated checks; 93 papers have both.

\begin{table}[htbp]
\centering
\caption{Whole-document semantic checks on nonempty search endpoints. A policy--seed endpoint is counted at each budget, even if the same text is reused at another checkpoint. Original choices are excluded from this validity denominator and remain zero-gain outcomes in the transfer analysis.}
\label{tab:endpoint_validity}
\begin{tabular}{lrrrr}
\toprule
Policy & Budget & Nonempty endpoints & Pass & Fail \\
\midrule
Adaptive & 64 & 254 & 242 (95.3\%) & 12 \\
Adaptive & 512 & 254 & 242 (95.3\%) & 12 \\
Random & 64 & 255 & 247 (96.9\%) & 8 \\
Random & 512 & 256 & 246 (96.1\%) & 10 \\
\bottomrule
\end{tabular}
\end{table}

There are 708 distinct nonempty composite endpoints at the checked budgets; 683 pass the semantic check and enter the frozen target inputs. Policy--seed counts in Table~\ref{tab:endpoint_validity} are larger because endpoints can recur. At 512, adaptive search retains the original for two of 256 policy--seed outcomes; random search retains it for none. At budget one, the original is retained for 137 adaptive and 146 random outcomes, so a small budget is not necessarily an active edit intervention.

\begin{table}[htbp]
\centering
\caption{Blind human audit of admitted items. Invalidity counts include uncertain judgments, although none occur in these three blind-audit strata. Confidence intervals (CIs) are exact Clopper--Pearson binomial intervals \citep{clopperpearson}. The adaptive subset is included within the overall endpoint row and is not an additional sample.}
\label{tab:human}
\resizebox{\linewidth}{!}{% Fit the table to the text block.
\begin{tabular}{lrrcc}
\toprule
Item & Audited & Invalid & Invalid rate [95\% CI] & One-sided upper bound \\
\midrule
Composite endpoint & 59 & 0 & 0.0\% [0.0, 6.1] & 4.95\% \\
\quad Adaptive-labeled subset & 19 & 0 & 0.0\% [0.0, 17.6] & 14.6\% \\
Presentation edit & 20 & 0 & 0.0\% [0.0, 16.8] & 13.9\% \\
Evidence removal & 20 & 6 & 30.0\% [11.9, 54.3] & 50.8\% \\
\bottomrule
\end{tabular}}
\end{table}

The endpoint sample includes 19 texts reached by adaptive search and 41 reached by random search; one text has both labels, so these counts sum to 60 over 59 distinct endpoints. The audit's overall endpoint bound therefore cannot be assigned to adaptive endpoints alone. Sampling is by paper, with one audited endpoint per selected paper. The 20 presentation edits are drawn uniformly from all admitted Study 2 single edits (seed 20261029) and come from 19 papers; the 20 removals come from 20 papers.

Of 1,002 primary-checker passes in the second-provider sample, 30 are disputed. The author adjudicates all 30, finding 13 invalid and one uncertain, an invalid-or-uncertain rate of 46.7\% $[28.3,65.7]$ for this selected disagreement set. The 13 invalid items comprise five changed claims, three changes in claim strength, four incoherent edits, and one mis-scoped removal. Twenty-two of the 30 also pass the invariant check, and hence can enter an admitted edit bank or endpoint; ten are invalid and one uncertain. Two disputed shams are judged invalid because they cut text mid-sentence. This is evidence about the editing pipeline as executed, even though the writer was instructed to delete whole paragraphs.

The disagreement adjudication covers 29 additional items after the 99-item audit, because one disputed item is already included there. Unlike the blind audit, the adjudicator may view the second checker's objection after an initial read. The observed 14 invalid-or-uncertain disputes among 1,002 primary passes are about 1.4\%; this is the fraction identified through that dispute pathway, not a complete error-rate estimate.

The cross-provider recheck covers 1,323 items and agrees with the primary checker on 86.6\%. Among items passed by the primary checker, the second provider passes 97.0\%; among items passed by the second provider, the primary checker passes 86.9\%. This asymmetry indicates that the primary checker is stricter in this sample, not that it is always correct. A 20-item challenge deliberately changes claims, hedges, attributions, or aims while preserving numbers. The primary checker rejects all 20. This tests those designed corruptions and does not estimate the miss rate for natural edits.

\appendixsection{Additional numerical results}
\label{app:extended}
Descriptive edit-type correlations pool edits and use 2,000 paper-cluster resamples. Their confidence intervals include zero, but they do not replace the mean of within-paper correlations in the primary analysis.
Tables~\ref{tab:kind}--\ref{tab:changes} provide the descriptive results underlying the main figures. All standardized changes use the fixed SDs from the 328 originals. Different intervention rows can have different paper cohorts. In particular, a paired removal-minus-sham estimate must be computed on its 93 paired papers rather than by subtracting marginal means on 102 and 184 papers.

Original-paper agreement also holds on the intervention cohorts themselves, not only on the 1,667-paper benchmark. In a post hoc check with 10,000 paper-bootstrap resamples, the Spearman rank correlation \citep{spearman1904} between \Surr{} and \Tone{} on the unedited papers is \CI{0.724}{0.637}{0.795} on the 198 fidelity papers and \CI{0.692}{0.586}{0.774} on the 128 Study~3 papers; for \Ttwo{} it is \CI{0.479}{0.356}{0.590} and \CI{0.417}{0.258}{0.555}. Thus, original-paper and edit-level agreement differ within the same paper cohorts. Agreement with \Ttwo{} is weaker than with \Tone{}, so the \Ttwo{} comparison tests a weaker premise.

\begin{table}[htbp]
\centering
\caption{Edit-type correlations, descriptive. These are pooled Spearman correlations \citep{spearman1904} with nominal 95\% paper-cluster bootstrap confidence intervals (CIs), rather than means of within-paper correlations.}
\label{tab:kind}
\begin{tabular}{lrrcc}
\toprule
Type & Papers & Edits & $T_1$ & $T_2$ \\
\midrule
Clarity & 199 & 1566 & \CI{0.040}{-0.022}{0.103} & \CI{0.024}{-0.033}{0.078} \\
Emphasis & 115 & 167 & \CI{-0.061}{-0.222}{0.110} & \CI{-0.093}{-0.272}{0.082} \\
Flow & 197 & 889 & \CI{0.011}{-0.055}{0.079} & \CI{0.009}{-0.068}{0.087} \\
Framing & 84 & 129 & \CI{0.094}{-0.098}{0.283} & \CI{-0.006}{-0.195}{0.178} \\
Structure & 193 & 572 & \CI{-0.067}{-0.149}{0.019} & \CI{0.056}{-0.021}{0.138} \\
\bottomrule
\end{tabular}
\end{table}

\begin{table}[htbp]
\centering
\caption{Adaptive minus random gain by maximum search budget. Each evaluator uses its own original-score standard deviation (SD). Only the two target contrasts at 512 enter the four-test family. All rows use 128 papers and nominal 95\% confidence intervals (CIs).}
\label{tab:budgets}
\begin{tabular}{rccc}
\toprule
Budget & $S$ & $T_1$ & $T_2$ \\
\midrule
1 & \CI{-0.004}{-0.006}{-0.002} & \CI{0.001}{-0.015}{0.018} & \CI{0.000}{-0.009}{0.010} \\
8 & \CI{-0.010}{-0.013}{-0.007} & \CI{0.001}{-0.020}{0.023} & \CI{-0.002}{-0.016}{0.012} \\
64 & \CI{0.010}{0.008}{0.012} & \CI{0.015}{-0.007}{0.039} & \CI{0.006}{-0.007}{0.019} \\
512 & \CI{0.011}{0.008}{0.013} & \CI{0.002}{-0.019}{0.024} & \CI{-0.001}{-0.014}{0.012} \\
\bottomrule
\end{tabular}
\end{table}

\begin{table}[htbp]
\centering
\caption{Post hoc single-edit selection from the complete variable-sized banks on the 198-paper fidelity cohort. Entries are target-score changes in target standard deviation (SD) units, with nominal 95\% confidence intervals (CIs). Oracle choices use the evaluated target's outcomes.}
\label{tab:headroom}
\begin{tabular}{lcc}
\toprule
Choice or contrast & $T_1$ & $T_2$ \\
\midrule
Uniform random expectation & \CI{-0.002}{-0.011}{0.008} & \CI{-0.002}{-0.009}{0.004} \\
Surrogate choice & \CI{-0.002}{-0.021}{0.018} & \CI{-0.006}{-0.017}{0.005} \\
Sibling choice & \CI{0.006}{-0.012}{0.022} & \CI{-0.005}{-0.015}{0.004} \\
Target oracle & \CI{0.152}{0.136}{0.169} & \CI{0.094}{0.083}{0.106} \\
Surrogate minus random & \CI{-0.000}{-0.016}{0.015} & \CI{-0.004}{-0.012}{0.004} \\
Sibling minus random & \CI{0.007}{-0.006}{0.020} & \CI{-0.003}{-0.010}{0.004} \\
Oracle minus random & \CI{0.153}{0.142}{0.165} & \CI{0.096}{0.088}{0.105} \\
\bottomrule
\end{tabular}
\end{table}

\begin{table}[htbp]
\centering
\caption{Changes relative to the original, in each evaluator's standard deviation (SD) units. Absolute single-edit changes average edits within each of 199 papers. Search and polish use 128 papers; search uses maximum budget 512 and averages seeds. Confidence intervals (CIs) are nominal 95\% paper-bootstrap intervals.}
\label{tab:changes}
\resizebox{\linewidth}{!}{% Fit the table to the text block.
\begin{tabular}{lccc}
\toprule
Quantity & $S$ & $T_1$ & $T_2$ \\
\midrule
Absolute single-edit change & \CI{0.010}{0.010}{0.011} & \CI{0.082}{0.076}{0.089} & \CI{0.050}{0.046}{0.055} \\
Adaptive minus original & \CI{0.064}{0.057}{0.071} & \CI{-0.025}{-0.055}{0.006} & \CI{0.006}{-0.014}{0.026} \\
Random minus original & \CI{0.053}{0.048}{0.058} & \CI{-0.027}{-0.055}{0.001} & \CI{0.006}{-0.010}{0.024} \\
Polish minus original & \CI{-0.018}{-0.029}{-0.008} & \CI{0.001}{-0.040}{0.042} & \CI{0.040}{0.012}{0.069} \\
\bottomrule
\end{tabular}}
\end{table}

For exploratory edit selection, each paper contributes one value for each chooser. Random selection is the average change across its admitted bank. Surrogate selection and sibling selection choose the edit with maximum change under that evaluator. Ties use the recorded edit order. The oracle uses the evaluated target itself to select the maximum. Papers can contain different numbers of admitted edits, so oracle gains depend on bank size. No additional target-scoring run is needed for this analysis, but its choice after observing the main results makes it post hoc.

\begin{figure}[t]
\centering
\includegraphics[width=\linewidth]{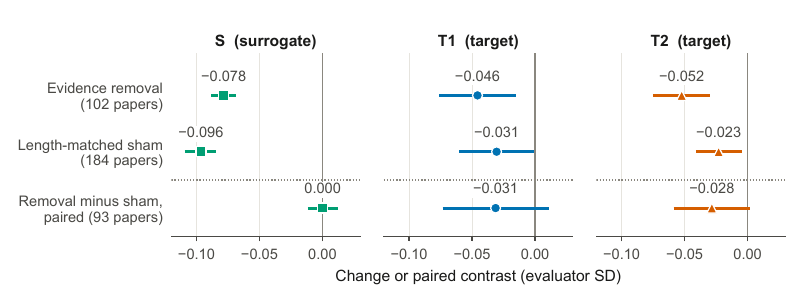}
\caption{The removal-minus-sham estimates are imprecise and confounded by coherence defects. One panel per evaluator. The first two rows show separate mean changes from the original for admitted evidence removals (102 papers) and length-matched shams (184 papers). The row below the dotted rule is the within-paper removal-minus-sham contrast on the 93 paired papers; it is not the difference between the first two means, which use different cohorts. Effects use each evaluator's original-score standard deviation (SD), with nominal 95\% paper-bootstrap confidence intervals (CIs) and point estimates above the marks. Human auditing finds coherence defects in six of 20 sampled removals, limiting a pure evidence-ablation interpretation.}
\label{fig:sensitivity}
\end{figure}

\begin{figure}[htbp]
\centering
\includegraphics[width=\linewidth]{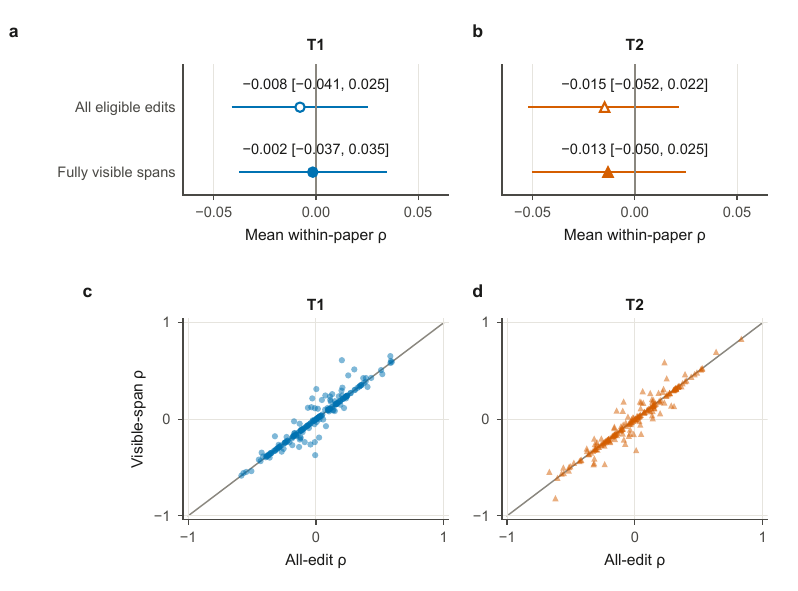}
\caption{Restricting edits to fully visible paragraphs leaves mean edit agreement near zero (post hoc). (a, b)~Mean within-paper Spearman correlation \citep{spearman1904} for \Tone{} and \Ttwo{}, using all eligible edits (open markers) or only paragraphs wholly retained by both targets and the surrogate's lexical window (filled markers). Both rows contain the same 198 papers: 3,321 edits before filtering and 3,072 after. Intervals are nominal 95\% paper-bootstrap confidence intervals; original estimates retain their original intervals. (c, d)~Each paper's correlation before and after filtering, with the identity line in gray. Overlapping points remain superimposed. The filter retains visible changed spans while leaving different surrounding model contexts and whole-document surrogate features intact; it is not an equalized-context experiment.}
\label{fig:visibility}
\end{figure}

\appendixsection{Evidence-removal diagnostic}
\label{app:removal}
\label{sec:sensitivity}
The evidence-removal analysis is pre-specified as descriptive. Only 93 papers have both an admitted removal and an admitted sham. On these paired papers, the removal-minus-sham contrast is \CI{0.000}{-0.012}{0.012} SD for \Surr, \CI{-0.031}{-0.073}{0.011} for \Tone, and \CI{-0.028}{-0.059}{0.002} for \Ttwo{} (Figure~\ref{fig:sensitivity}). Both target point estimates are negative, but the intervals include zero. The surrogate's near-zero contrast is consistent with a response to deletion-related features rather than the distinction between evidence and neutral text; it does not prove that the model is universally blind to evidence.

Across the 102 admitted removals, the mean absolute response is 0.125 SD for \Tone, compared with 0.082 for a single presentation edit across 199 papers. These magnitudes are descriptive and come from different cohorts. They indicate that a wording edit can move this target on a scale comparable to the implemented deletion. They do not establish that the model values wording more than scientific evidence. Excluding the five audited-invalid removals that enter the paired analysis leaves 88 pairs: \Tone{} gives $-0.031$ SD $[-0.075,0.013]$, and \Ttwo{} $-0.033$ $[-0.065,-0.002]$. This post hoc exclusion leaves unaudited pairs unchanged and does not repair the remaining deletions or turn the deletion analysis into a confirmatory result.

\emph{32-bit rescoring (post hoc).} We rescored the 102 removals and 184 shams in 32-bit on the RTX 4090, as for the edit-fidelity set (Supplementary Section~\ref{app:reliability}); changes are in SD units of the 200 Study 2 originals' 32-bit scores. The removal-minus-sham contrast is \CI{-0.014}{-0.051}{0.023} for \Tone{} and \CI{-0.019}{-0.047}{0.010} for \Ttwo{} on the 93 paired papers, and \CI{-0.018}{-0.059}{0.020} and \CI{-0.022}{-0.051}{0.010} on the 88 pairs without audited-invalid removals. The \Ttwo{} interval that excluded zero after the exclusion in bfloat16 therefore includes zero at full precision. Removals and shams move both targets in the same direction (\Tone: \CI{-0.046}{-0.080}{-0.014} and \CI{-0.040}{-0.063}{-0.017}; \Ttwo: \CI{-0.053}{-0.074}{-0.033} and \CI{-0.023}{-0.042}{-0.004}).

\appendixsection{Post hoc visibility, search, and fixed-bank analyses}
\label{app:posthoc}
These descriptive analyses reuse the frozen texts and classifier scores. Their analysis choices were fixed locally after the primary results were known and before these analyses were computed; they are not a public preregistration or an extension of the four-test family. Sections S6.1--S6.3 use saved target scores without further inference or model training; Section~\ref{app:reliability} reports separate post hoc rescoring. The post hoc intervals use 10,000 paper-bootstrap resamples, with seed 20261030 in Sections S6.1--S6.3 and seed 0 in Section~\ref{app:reliability}, except the 32-bit fixed-bank repeat (seed 20261030); estimates from the primary analysis retain their original bootstrap intervals.

\subsection{Visibility reconstruction}

For every single edit, we verify exact equality outside the replaced paragraph and locate the complete original and replacement spans. We reconstruct each target prompt with its saved tokenizer and original domain-specific instructions, including decoding a truncated token prefix before inserting it into the prompt. This reproduces both the saved truncation flag and the total prompt length for all 3,651 originals and single-edit versions in each target, or 7,302 prompt reconstructions. The text-body budgets are 24,375 tokens for \Tone{} and 24,370 for \Ttwo{} within the 24,480-token scoring limit.

A common-visible edit must have both complete paragraph spans inside each target's retained text and inside the surrogate's first 30,000 characters. All 3,323 admitted single edits meet the target condition. The surrogate-window-only and common-visible restrictions therefore coincide: 3,074 edits remain across 199 papers, of which 3,072 belong to the original 198-paper fidelity cohort. Eligibility is still at least five edits and nonconstant scores. Table~\ref{tab:visibility} also reports a stricter subset whose complete original and edited documents fit every input limit; its six papers provide weak precision. None of these filters removes the surrogate's whole-document structural channel or gives every model an identical context.

\paragraph{One-pass polishing}
The filtered one-pass polish changes \Tone{} by \CI{0.001}{-0.040}{0.042} SD, and \Ttwo{} by \CI{0.040}{0.012}{0.069}. It lowers \Surr{} by \CI{-0.018}{-0.029}{-0.008} SD. These are descriptive contrasts, outside the four-test family. \Ttwo's interval excludes zero, while \Tone's is wide and contains \Ttwo's estimate; we did not test the difference between targets. The result does not establish that polishing improves scientific merit.

\subsection{Search outcomes and realized consumption}
\begin{figure}[htbp]
\centering
\includegraphics[width=\linewidth]{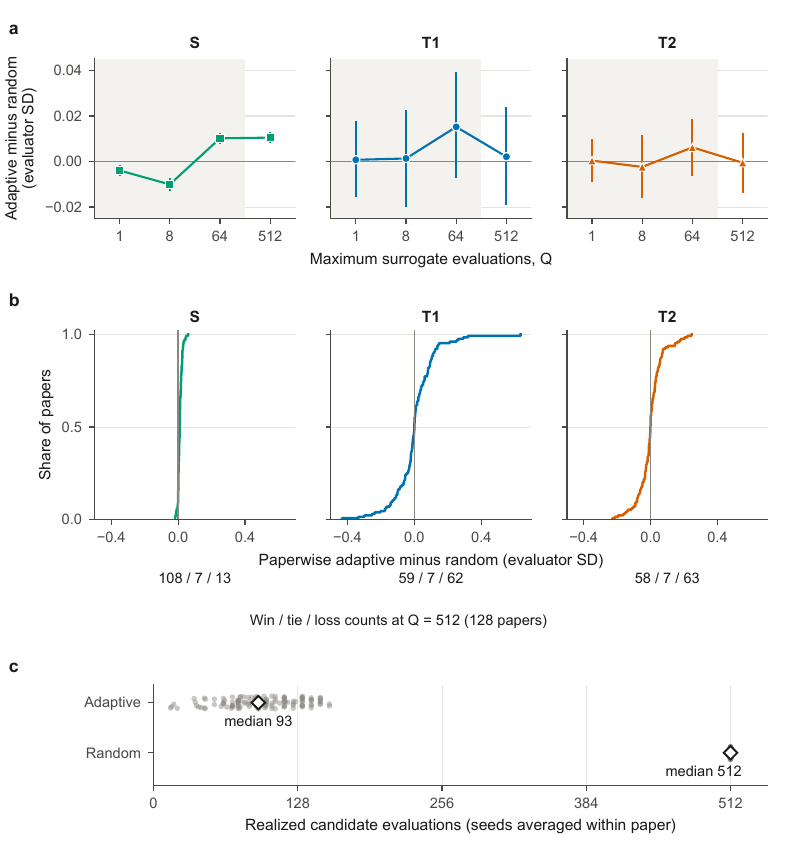}
\caption{Adaptive search has a small average source advantage, variable target effects, and unequal realized consumption. (a)~Adaptive minus best-of-$Q$ random gain at each maximum allowance, with both policies selecting according to \Surr; the original remains eligible. Only the two target contrasts at $Q=512$ belong to the four-test family. Shading marks exploratory budgets; budgets $Q=1$ and $Q=8$ lack complete endpoint checks. (b)~Empirical cumulative distributions of the 128 paperwise contrasts at 512, with win/tie/loss counts below. Columns show \Surr, \Tone, and \Ttwo{} in their own original-score standard deviation (SD) units on shared axes. (c)~Actual candidate evaluations recovered by exact replay, with seeds averaged within each paper; points are vertically offset for visibility and diamonds mark medians. Intervals in (a) are the original nominal 95\% paper-bootstrap confidence intervals; (b, c) are post hoc descriptions of observed variation. Invalid checked endpoints contribute zero gain.}
\label{fig:transfer}
\end{figure}
\begin{figure}[htbp]
\centering
\includegraphics[width=\linewidth]{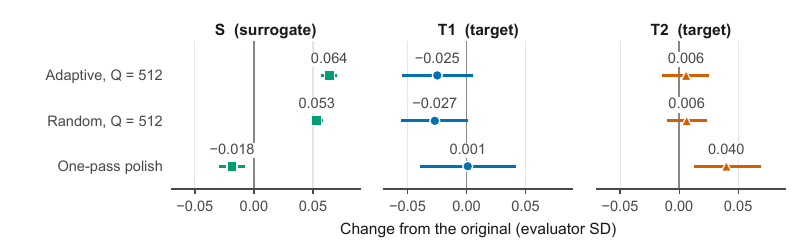}
\caption{Original-relative controls retain the small source-objective advantage and the uncertainty of target changes. Changes from the original for adaptive search, random search, and filtered one-pass polish. Columns show \Surr, \Tone, and \Ttwo{}, each standardized by its own original-score standard deviation (SD). Intervals are the original nominal 95\% paper-bootstrap confidence intervals, averaging the two search seeds within each of 128 papers. Invalid checked endpoints count as zero gain. The polishing contrasts are descriptive.}
\label{fig:search_controls}
\end{figure}
At 512, we average the two seeds within each paper and plot the distribution of adaptive-minus-random changes (Figure~\ref{fig:transfer}b). Adaptive search wins and loses against random search on comparable numbers of papers for both targets, while it usually wins on \Surr{} (Table~\ref{tab:search_paperwise}). Exact ties are counted before rounding. Invalid or original endpoints retain zero gain as in the primary analysis. To recover evaluation counts, we replay the original search with the saved surrogate weights, source texts, titles, admitted banks, policy seeds, feasibility rules, and original-outside-budget convention. Every checkpoint reproduces the exact chosen edit indices and its saved surrogate score to within~$10^{-12}$. The replay covers 128 papers, two policies, two seeds, and four checkpoints; it uses only the central-processing-unit surrogate, not either target.

Both adaptive seeds have identical realized counts within each paper. Counts range from 15 to 156, with median 93 and interquartile range 75--120.25; all random runs consume 512. Thus, checkpoint 512 is a common maximum allowance, not 512 evaluations actually used by both policies. 

\begin{samepage}
Figure~\ref{fig:search_controls} supplies the original-relative controls, including polishing. The descriptive paperwise distributions do not establish a new average search benefit or a causal explanation for individual gains.
\end{samepage}

\subsection{Equal-size selection and bank-size dependence}
\begin{figure}[htbp]
\centering
\includegraphics[width=\linewidth]{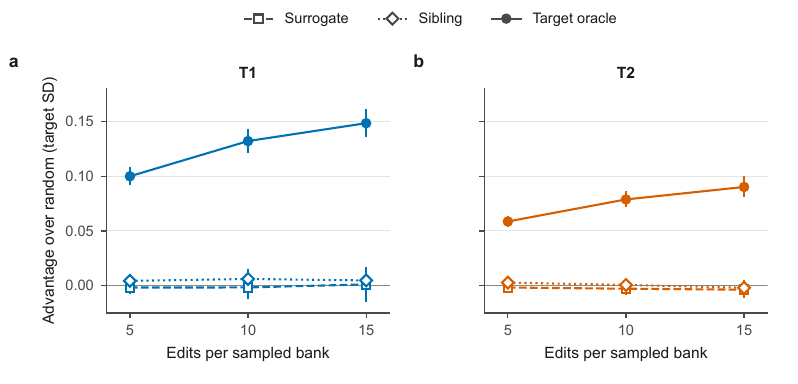}
\caption{Larger banks increase oracle advantage without a clear proxy advantage on the fixed cohort (post hoc). Panels show \Tone{} and \Ttwo{} on the same 160 papers with at least fifteen admitted edits. Lines compare surrogate, sibling, and outcome-informed oracle choice against the uniform-random expectation of the same sampled bank, at five, ten, and fifteen edits. Each estimate averages 1,000 draws within paper before a 10,000-resample paper bootstrap; intervals are nominal 95\% confidence intervals. Units are target original-score standard deviations (SDs). The oracle's positive advantage reflects selection by construction. Fixed cohort membership prevents changing paper composition from explaining its increase.}
\label{fig:bank_size}
\end{figure}
Each original fidelity paper contributes equally. We draw 1,000 banks independently for each paper and bank size, sampling edits without replacement within each bank, using seed 20261031 combined deterministically with the paper identifier and bank size. Within each draw the target change under a chooser is compared with the mean change across that exact sampled bank. Surrogate and sibling choosers use only their respective scores; the oracle uses the evaluated target itself. Ties use recorded edit order. Oracle-minus-choice regret is the score difference between the oracle's edit and the proxy's edit, averaged within paper. We average draws within each paper before bootstrap resampling, so draws and edits do not become independent sampling units.

All 198 fidelity papers support banks of five. The 160 papers with at least fifteen admitted edits form the fixed cohort for sizes five, ten, and fifteen. Figure~\ref{fig:bank_size} holds that cohort constant. Conditional on the observed banks, the aggregate Monte Carlo standard error for each displayed mean is at most 0.00021 target SD, smaller than the displayed paper-bootstrap uncertainty; the latter intervals do not include this simulation error. The median original bank size is seventeen. Results for the complete, variable-sized banks remain in Table~\ref{tab:headroom} and must not be compared with the five-edit estimates as if only the chooser changed.

The score lost by choosing with a proxy rather than the evaluated target also varies by paper (Figure~\ref{fig:headroom}c, d); in 32-bit, its mean for surrogate choice is 0.078 SD on \Tone{} and 0.054 SD on \Ttwo{}. Increasing the sampled bank size from five to fifteen on the same 160 papers increases the oracle advantage, while all surrogate- and sibling-choice intervals still span zero (Supplementary Figure~\ref{fig:bank_size}). Holding bank size fixed therefore preserves the descriptive finding that these proxies add little over random choice. It does not test a surrogate trained specifically to predict edit responses.

The oracle advantage and oracle-minus-proxy regret are nonnegative by their definitions. They quantify response variation in finite banks, rather than evidence that any edit improves scientific quality, or a performance guarantee for new banks. A proxy advantage near zero means that it does little better on average than random selection from these same available alternatives.

\posthoctablebarrier
\begin{table}[htbp]
\centering
\small
\setlength{\tabcolsep}{4pt}
\caption{Visibility-restricted mean within-paper correlations. All rows use the original fidelity-cohort papers that still meet the five-edit and nonconstant-score rules. The common-visible filter equals the surrogate-window-only filter because all edited paragraphs are visible to both targets. Confidence intervals (CIs) are nominal 95\% paper-bootstrap intervals; original intervals are retained in the first row. The six-paper complete-document subset is imprecise and does not establish an equal-context result.}
\label{tab:visibility}
\begin{tabular}{lrrcc}
\toprule
Subset & Papers & Edits & $S,T_1$ [95\% CI] & $S,T_2$ [95\% CI] \\
\midrule
All eligible edits & 198 & 3321 & \CI{-0.008}{-0.041}{0.025} & \CI{-0.015}{-0.052}{0.022} \\
Common-visible spans & 198 & 3072 & \CI{-0.002}{-0.037}{0.035} & \CI{-0.013}{-0.050}{0.025} \\
Complete documents fit & 6 & 120 & \CI{0.044}{-0.153}{0.215} & \CI{0.134}{-0.057}{0.334} \\
\bottomrule
\end{tabular}
\end{table}
\begin{table}[htbp]
\centering
\small
\setlength{\tabcolsep}{4pt}
\caption{Paperwise adaptive-minus-random changes at the maximum allowance of 512. Means and nominal 95\% confidence intervals (CIs) retain the original analysis; counts classify the 128 paperwise seed-averaged differences before rounding. Invalid or original endpoints retain zero gain. Units are each evaluator's original-score standard deviation (SD).}
\label{tab:search_paperwise}
\begin{tabular}{lcrrr}
\toprule
Evaluator & Mean [95\% CI] & Win & Tie & Loss \\
\midrule
\Surr & \CI{0.011}{0.008}{0.013} & 108 & 7 & 13 \\
\Tone & \CI{0.002}{-0.019}{0.024} & 59 & 7 & 62 \\
\Ttwo & \CI{-0.001}{-0.014}{0.012} & 58 & 7 & 63 \\
\bottomrule
\end{tabular}
\end{table}
\begin{table}[htbp]
\centering
\small
\setlength{\tabcolsep}{4pt}
\caption{Realized candidate evaluations under the common maximum allowance of 512, recovered by exact surrogate replay. Both seeds have identical counts within every paper, so these paperwise summaries also describe the 256 runs per policy. The original paper is scored once outside this count. Interquartile ranges summarize observed dispersion, not confidence intervals.}
\label{tab:search_consumption}
\begin{tabular}{lrrrr}
\toprule
Policy & Papers & Median & Interquartile range & Range \\
\midrule
Adaptive & 128 & 93 & 75--120.25 & 15--156 \\
Random & 128 & 512 & 512--512 & 512--512 \\
\bottomrule
\end{tabular}
\end{table}
\begin{table}[htbp]
\centering
\small
\setlength{\tabcolsep}{4pt}
\caption{Choice advantage over the random expectation of the same five-edit bank, in target standard deviation (SD) units. All 198 fidelity papers contribute equally after averaging 1,000 draws within paper. Confidence intervals (CIs) are nominal 95\% paper-bootstrap intervals conditional on the observed banks and exclude Monte Carlo error. The oracle uses target outcomes, so its advantage is nonnegative by construction.}
\label{tab:fixed_bank}
\begin{tabular}{lcc}
\toprule
Chooser & $T_1$ advantage [95\% CI] & $T_2$ advantage [95\% CI] \\
\midrule
Surrogate choice & \CI{-0.002}{-0.008}{0.003} & \CI{-0.002}{-0.005}{0.001} \\
Sibling choice & \CI{0.005}{-0.001}{0.010} & \CI{0.002}{-0.002}{0.005} \\
Target oracle & \CI{0.101}{0.094}{0.108} & \CI{0.061}{0.056}{0.066} \\
\bottomrule
\end{tabular}
\end{table}
\begin{table}[htbp]
\centering
\small
\setlength{\tabcolsep}{4pt}
\caption{Bank-size sensitivity on the same 160 papers with at least fifteen admitted edits. Every entry is a choice-minus-random advantage in target standard deviation (SD) units, averaged over 1,000 draws per paper. Confidence intervals (CIs) are nominal 95\% paper-bootstrap intervals. Holding the cohort constant avoids confounding bank size with paper composition.}
\label{tab:bank_size}
\begin{tabular}{rlcc}
\toprule
Bank size & Chooser & $T_1$ [95\% CI] & $T_2$ [95\% CI] \\
\midrule
5 & Surrogate & \CI{-0.002}{-0.008}{0.004} & \CI{-0.002}{-0.005}{0.001} \\
5 & Sibling & \CI{0.004}{-0.001}{0.010} & \CI{0.003}{-0.001}{0.006} \\
5 & Target oracle & \CI{0.100}{0.092}{0.108} & \CI{0.059}{0.054}{0.064} \\
10 & Surrogate & \CI{-0.002}{-0.012}{0.008} & \CI{-0.003}{-0.008}{0.002} \\
10 & Sibling & \CI{0.006}{-0.003}{0.015} & \CI{0.001}{-0.005}{0.006} \\
10 & Target oracle & \CI{0.132}{0.121}{0.143} & \CI{0.079}{0.072}{0.086} \\
15 & Surrogate & \CI{0.001}{-0.015}{0.015} & \CI{-0.004}{-0.011}{0.004} \\
15 & Sibling & \CI{0.005}{-0.007}{0.017} & \CI{-0.002}{-0.009}{0.005} \\
15 & Target oracle & \CI{0.148}{0.136}{0.161} & \CI{0.090}{0.081}{0.100} \\
\bottomrule
\end{tabular}
\end{table}

\subsection{Reliability of target edit responses}
\label{app:reliability}
We rescored a seeded sample of 25 Study 2 fidelity papers (seed 20261101) on an NVIDIA RTX 5090 with the same pipeline, software versions and bfloat16 forward pass. The sample covers the 25 originals, their 438 admitted single edits, and five whitespace-only null edits per paper; each null edit widens one sentence gap from one space to two. For all 463 originals and real edits shared with the RTX 4090 run, token counts match. Repeated scoring of ten originals on the same GPU is bit-identical. Edit-response correlations use the estimand and bootstrap of Equation~\ref{eq:fidelity}, with RTX 4090 and RTX 5090 changes in place of the two evaluators (Table~\ref{tab:reliability}).

\begin{table}[htbp]
\centering
\small
\setlength{\tabcolsep}{3pt}
\caption{Reliability of target edit responses on 25 Study 2 papers (post hoc). RTX 5090 rescoring is compared with the RTX 4090 scores used throughout the paper; a further 32-bit run on the RTX 4090 separates precision from hardware. Original-score correlations are Pearson correlations across the 25 papers; edit correlations are mean within-paper Spearman correlations \citep{spearman1904} with nominal 95\% paper-bootstrap confidence intervals (CIs). Absolute changes pool the 438 real edits or 125 null edits, in each target's original-score standard deviation (SD); null edits widen one sentence gap and change no word.}
\label{tab:reliability}
\begin{tabular}{llcc}
\toprule
Quantity & Scoring & $T_1$ & $T_2$ \\
\midrule
Original scores, $r$ with RTX 4090 & bfloat16 & 0.9985 & 0.9993 \\
Edit changes vs.\ RTX 4090 & bfloat16 & \CI{0.681}{0.604}{0.755} & \CI{0.658}{0.603}{0.712} \\
 & 32-bit & \CI{0.379}{0.266}{0.483} & \CI{0.455}{0.366}{0.538} \\
Edit changes vs.\ 4090 32-bit & 32-bit & \CI{0.9997}{0.9994}{1.000} & \CI{0.9998}{0.9995}{1.000} \\
Mean $|\Delta|$ (SD), real / null & bfloat16 & 0.078 / 0.052 & 0.055 / 0.034 \\
 & 32-bit (4090) & 0.055 / 0.026 & 0.030 / 0.015 \\
Agreement with $S$, RTX 4090 & bfloat16 & \CI{0.019}{-0.085}{0.116} & \CI{-0.020}{-0.132}{0.087} \\
Agreement with $S$, RTX 5090 & bfloat16 & \CI{0.016}{-0.072}{0.098} & \CI{-0.028}{-0.108}{0.050} \\
 & 32-bit & \CI{0.022}{-0.082}{0.126} & \CI{0.029}{-0.071}{0.131} \\
\bottomrule
\end{tabular}
\end{table}

Original-paper scores agree closely across GPUs: Pearson correlations are 0.9985 for \Tone{} and 0.9993 for \Ttwo{}, and the mean absolute probability difference is 0.009 for both. Edit changes reproduce less well because they are small, with mean absolute sizes on the RTX 5090 of 0.078 SD for \Tone{} and 0.055 SD for \Ttwo. The within-paper correlation between the two GPUs' edit changes is \CI{0.681}{0.604}{0.755} for \Tone{} and \CI{0.658}{0.603}{0.712} for \Ttwo. Whitespace-only edits move \Tone{} by 0.052 SD and \Ttwo{} by 0.034 SD, about two thirds of the real-edit movement. They move \Surr{} by 0.002 SD. Each whitespace edit adds one token, so this checks formatting sensitivity rather than repeated scoring of identical inputs.

For a descriptive sensitivity calculation, assume independent classical measurement errors, no measurement error in \Surr, and that the cross-GPU correlations estimate target reliability. Under these assumptions, agreement with each target is attenuated by the square root of its reliability. Dividing the primary upper limits by these factors, using unrounded estimates, gives 0.031 and 0.027. Dividing the sibling upper limit of 0.075 by the square root of the product of both reliabilities gives 0.112. This scaling is approximate for averaged rank correlations and rests on 25 papers; using the lower reliability limits gives 0.033, 0.028 and 0.124. The scaled endpoints are not noise-corrected confidence bounds: the error model is unverified and reliability uncertainty is not propagated. On this subset, agreement between \Surr{} and each target stays near zero with the RTX 5090 scores: \CI{0.016}{-0.072}{0.098} for \Tone{} and \CI{-0.028}{-0.108}{0.050} for \Ttwo.

Scoring the whole model in 32-bit floating point changes the targets more than changing GPU. Original scores shift by a mean absolute probability difference of 0.089 for \Tone{} and 0.175 for \Ttwo{}. Edit changes correlate with the RTX 4090 changes at \CI{0.379}{0.266}{0.483} and \CI{0.455}{0.366}{0.538}. Agreement with \Surr{} stays near zero: \CI{0.022}{-0.082}{0.126} and \CI{0.029}{-0.071}{0.131}. To separate precision from hardware, we repeat the 32-bit scoring of all 588 inputs on the RTX 4090. The 32-bit edit changes of the two GPUs correlate at \CI{0.9997}{0.9994}{1.000} for \Tone{} and \CI{0.9998}{0.9995}{1.000} for \Ttwo, and the RTX 4090 32-bit changes correlate with its own bfloat16 changes at 0.380 and 0.455, as the RTX 5090 32-bit changes do. The lower cross-precision reliability therefore reflects bfloat16 rounding, not hardware: the bfloat16 changes carry a component, partly shared across GPUs, that the 32-bit model does not produce. In 32-bit, whitespace-only edits still move \Tone{} by 0.026 SD and \Ttwo{} by 0.015 SD, about half of the real-edit movement (0.055 and 0.030 SD). On these 25 papers, 32-bit target agreement is \CI{-0.043}{-0.176}{0.084}.

On the 1,667-paper benchmark, a 32-bit pass on the RTX 4090 gives AUCs of \CI{0.713}{0.688}{0.736} for \Tone{} and \CI{0.666}{0.640}{0.691} for \Ttwo, against 0.720 and 0.670 in bfloat16. Paper rankings change little across precisions (Spearman 0.909 for \Tone{} and 0.983 for \Ttwo), although 32-bit lowers the mean predicted acceptance from 0.499 to 0.477 and from 0.573 to 0.409. Original-paper agreement persists in 32-bit: \Surr's Spearman correlation is 0.680 with \Tone{} and 0.509 with \Ttwo{} (0.731 and 0.547 in bfloat16), and the two targets correlate at 0.625 (0.626).

\emph{Primary estimand in 32-bit.} We rescore the 200 Study 2 originals and 3,323 admitted single edits with both targets in 32-bit on the RTX 4090. With \Surr's changes unchanged, the mean within-paper agreement on the 198 fidelity papers is \CI{-0.008}{-0.047}{0.031} for \Tone{} and \CI{-0.021}{-0.059}{0.016} for \Ttwo, against \CI{-0.008}{-0.041}{0.025} and \CI{-0.015}{-0.052}{0.022} in bfloat16 (Table~\ref{tab:confirmatory}). The targets agree at \CI{0.047}{0.004}{0.090} in 32-bit, against 0.033 in bfloat16. Within papers, the bfloat16 edit changes correlate with the 32-bit changes at \CI{0.284}{0.245}{0.321} for \Tone{} and \CI{0.496}{0.461}{0.530} for \Ttwo. Averaged within papers, the 32-bit edit changes are 0.056 and 0.040 of each target's 32-bit SD across the 200 originals (0.242 and 0.257), against 0.082 and 0.050 in bfloat16. Because 32-bit scores reproduce across GPUs, these estimates carry little of the rounding error that the scaled endpoints above approximate, and they supersede that scaling. Supplementary Section~\ref{app:removal} gives the evidence-removal contrast in 32-bit.

\emph{Search in 32-bit.} We rescore the 128 Study 3 originals, the 683 admitted search endpoints, and the 128 polished versions with both targets in 32-bit on the RTX 4090, and repeat the pre-specified contrast with the frozen endpoints. In units of each target's 32-bit SD across the 328 study originals (0.244 and 0.256), adaptive minus random gain at $Q=512$ is \CI{-0.015}{-0.033}{0.001} for \Tone{} and \CI{-0.006}{-0.022}{0.009} for \Ttwo, against 0.002 and $-0.001$ in bfloat16. Adaptive gains over the original are \CI{-0.011}{-0.034}{0.011} and \CI{-0.005}{-0.028}{0.016}, and random gains \CI{0.004}{-0.014}{0.022} and \CI{0.001}{-0.015}{0.018}.

\emph{Float16, a proxy for the release recipe.} The release serves its 32-bit weights through vLLM, whose default runs them in float16. To approximate that recipe, we rescore the 200 Study 2 originals and 3,323 admitted single edits with a float16 forward pass in Transformers on the RTX 4090, reading the decision in 32-bit as before; no score is non-finite. Within papers, the float16 edit changes correlate with the 32-bit changes at \CI{0.978}{0.974}{0.982} for \Tone{} and \CI{0.976}{0.972}{0.979} for \Ttwo, but with the bfloat16 changes at only \CI{0.281}{0.242}{0.319} and \CI{0.499}{0.464}{0.534}. With float16 targets, edit fidelity on the 198 fidelity papers is \CI{-0.004}{-0.043}{0.036} for \Tone{} and \CI{-0.026}{-0.063}{0.012} for \Ttwo, and the targets agree at \CI{0.045}{0.003}{0.087}, as in 32-bit. On the 1,667-paper benchmark, float16 gives AUCs of 0.713 for \Tone{} and 0.666 for \Ttwo{}. Each point estimate differs from 32-bit by less than 0.0002, and both paired-bootstrap CIs include zero. Its paper scores have Spearman correlations of at least 0.9999 with the 32-bit scores. vLLM's kernels may round differently, so this pass approximates the recipe rather than reproducing it.

\emph{Prompt swap.} The targets differ in weights and in prompt. To separate the two, we rescore the 200 Study 2 originals and 3,323 admitted single edits with \Tone{} in float16 under \Ttwo's arXiv preamble instead of its own ICLR preamble (Supplementary Section~\ref{app:surrogate}). On the 198 fidelity papers, \Tone's edit changes under the swapped prompt agree with \Ttwo's at \CI{0.067}{0.027}{0.107}, against \CI{0.045}{0.003}{0.087} under its own prompt, and with \Surr's at \CI{-0.016}{-0.054}{0.022}. The prompt therefore does not account for the targets' disagreement. The swap leaves \Tone's paper rankings almost unchanged (Spearman 0.986 on the 200 originals; 0.564 with \Ttwo, against 0.562), yet its edit changes under the two prompts agree at only \CI{0.537}{0.497}{0.575}: within one set of weights, a change of instructions preserves paper rankings and reorders the edits.

\emph{Fixed banks in 32-bit.} We repeat the five-edit bank analysis of Section~\ref{sec:headroom} with the 32-bit target changes, using the same banks, draws, and seeds; the bfloat16 run reproduces the reported values exactly. Choosing with the evaluated target gives \CI{0.076}{0.071}{0.082} SD for \Tone{} and \CI{0.052}{0.047}{0.058} for \Ttwo, against 0.101 and 0.061 in bfloat16. Surrogate choice gives \CI{-0.002}{-0.005}{0.002} and \CI{-0.001}{-0.004}{0.001}, and sibling choice stays within 0.005 SD of random choice. On the 160-paper cohort, the 32-bit oracle advantage rises from 0.076 to 0.120 for \Tone{} and from 0.049 to 0.077 for \Ttwo{} as banks grow from five to fifteen edits. An oracle that chooses with bfloat16 scores but is evaluated in 32-bit keeps \CI{0.024}{0.020}{0.029} and \CI{0.030}{0.026}{0.035} SD: the reduced-precision oracle partly selects rounding error, while the 32-bit oracle advantage stays well above the proxies'.

\emph{Whitespace twins in 32-bit.} To measure how reproducible the 32-bit edit responses are, we pair each admitted single edit of a seeded 60 of the 198 fidelity papers (seed 20261102; 1,000 edits) with a twin: the same edited text with one sentence gap widened from one space to two. The gap is drawn among sentence boundaries between characters 2,000 and 30,000 and at least 200 characters from the changed span, so the twin leaves the edit intact. Both targets score the twins in 32-bit on the RTX 4090; \Surr{} scores them on the CPU. The within-paper correlation between each edit's change and its twin's change is \CI{0.760}{0.713}{0.804} for \Tone, \CI{0.743}{0.692}{0.783} for \Ttwo, and \CI{0.960}{0.949}{0.970} for \Surr, although the added space moves the targets by 0.032 and 0.020 SD per edit, against edit changes of 0.058 and 0.036 SD (0.78 and 0.52 against 1.40 and 0.93 points of acceptance score). If these correlations serve as reliability proxies under independent-error assumptions, the classical attenuation relation $\sqrt{r_a r_b}$ \citep{spearman1904} gives illustrative ceilings of 0.854 for \Surr--\Tone, 0.845 for \Surr--\Ttwo, and 0.752 for \Tone--\Ttwo{} (0.823, 0.811, and 0.703 at the lower interval limits). On the same papers the observed agreements are \CI{-0.037}{-0.109}{0.037}, \CI{-0.030}{-0.107}{0.048}, and \CI{0.036}{-0.034}{0.105}. This comparison uses one kind of meaning-null change; other perturbations could show lower reliability, and reproducible responses need not track content.

\emph{Above-floor edits, a distilled surrogate, and topic (post hoc, saved scores).} Taking the 95th percentile of the added-space change over the 1,000 twins as a floor (0.093 SD for \Tone{} and 0.071 for \Ttwo{} in 32-bit), 672 edits on 157 papers exceed it for \Tone{} and 558 on 127 papers for \Ttwo; on the 62 and 50 papers with at least five such edits (417 and 396 edits), their agreement with \Surr{} is \CI{0.050}{-0.068}{0.166} and \CI{-0.045}{-0.166}{0.078}. With the median as floor, it is \CI{-0.016}{-0.063}{0.032} on 181 papers and \CI{-0.039}{-0.086}{0.006} on 177, and the targets agree at \CI{0.052}{-0.011}{0.113} on 158; only three papers have five edits above both 95th-percentile floors. A ridge model on \Surr's input features, fitted to each target's 32-bit scores of the 1,339 benchmark papers outside the 328 study originals (penalty by five-fold cross-validation), ranks the 198 held-out fidelity originals at 0.645 with \Tone{} and 0.701 with \Ttwo, yet its edit changes agree with the targets' 32-bit changes at only \CI{-0.002}{-0.043}{0.039} and \CI{0.033}{-0.003}{0.071}. Clustering the titles and abstracts of the 1,667 benchmark papers into 10, 20, or 40 topics (k-means on a separate TF--IDF) leaves \Surr's agreement with \Tone{} at 0.72 within topics (0.73 pooled) and at 0.57--0.58 when the structure counts are also residualized (0.60 pooled); for \Ttwo{} the values are 0.57--0.58 (0.55) and 0.24 (0.23). Topics explain 2--8\% of the rank variance of each score. The clusters approximate topic; the venue's own subject areas were not used.
\FloatBarrier

\subsection{Additional post hoc analyses}
\label{app:postreview}
These descriptive analyses were run after the four pre-specified tests and are outside that family. Most reuse saved scores; the positive control scores newly damaged texts with the paper's pipeline, and the dense surrogate trains a new classifier, with both configurations fixed before their outcomes were examined. Table~\ref{tab:postreview} collects them. Intervals are nominal 95\% paper-bootstrap intervals (seed 0; 10,000 resamples for agreement on the 25-paper subset, 2,000 for the pooled and between-paper correlations).

\emph{Target agreement and reliability.} On the 25 rescored papers, the two targets' within-paper edit agreement stays near zero under every scoring: RTX 4090 bfloat16, RTX 5090 bfloat16, and RTX 5090 32-bit (Table~\ref{tab:postreview}). On the 125 whitespace-only edits of the same papers, their agreement is \CI{0.136}{-0.060}{0.320}; \Surr's whitespace changes are identical within each paper, so its null-edit agreement is undefined. Real edits vary more than whitespace edits: the variance of real-edit changes exceeds the null-edit variance by 59\% of the real-edit variance for \Tone{} and 65\% for \Ttwo. Replacing the cross-GPU reliabilities of Section~\ref{app:reliability} with the cross-precision values (0.379 and 0.455) scales the primary upper limits to 0.041 and 0.032 for \Surr{} and to 0.180 between the targets, under the same unverified error model; the direct 32-bit estimates in Section~\ref{app:reliability} supersede this scaling. On the 198 fidelity papers, the targets' original-paper scores correlate at \CI{0.603}{0.501}{0.688} (Spearman; seed 20261031).

\emph{What moves each scorer.} Within papers, we regress each scorer's standardized edit change on the character-count change, its absolute value, the paragraph's relative position, whether the paragraph lies outside the surrogate's window, and the edit type, after subtracting paper means. These descriptors explain 1.0\% of the within-paper variance of \Surr's changes and 0.2\% and 0.4\% for the bfloat16 changes of \Tone{} and \Ttwo{} (0.8\% and 0.4\% in 32-bit); for absolute changes, 16.0\%, 3.8\%, and 2.2\%. The absolute size of an edit's character change correlates with the absolute score change at 0.32 for \Surr{} and 0.18 and 0.15 for the targets. No admitted single edit truncates a target input.

\emph{Levels of analysis.} On the 1,667 benchmark papers, \Surr's Spearman correlation with \Tone{} is 0.731, and 0.602 after ranking out the structure counts (length, figures, tables, equations, headings, citations, title and abstract length) from both scores. With \Ttwo{} it falls from 0.547 to 0.226; the structure-only score alone correlates at 0.511 with \Tone{} and 0.579 with \Ttwo. \Ttwo's agreement with \Surr{} on existing papers therefore runs largely through document structure that the edits hold fixed.

\emph{Estimand comparison and heterogeneity.} \citet{gainagreement} report between-paper Pearson correlations of whole-paper gains. Pooling all 3,321 fidelity edits gives Pearson correlations of 0.018, 0.024, and 0.025 for \Surr--\Tone, \Surr--\Ttwo, and \Tone--\Ttwo; correlating each paper's mean change across the 198 papers gives 0.099, 0.107, and 0.003, all with intervals spanning zero (Table~\ref{tab:postreview}). The spread of per-paper surrogate--target correlations matches a within-paper permutation null (SD 0.244 against 0.256 for \Tone{}; 0.264 against 0.257 for \Ttwo{}; 2,000 permutations), whereas the target--target correlations vary more than the null predicts (0.301 against 0.256; permutation $p<0.001$): the two targets agree on some papers and disagree on others.

\emph{Positive control.} To check that the within-paper estimand can register agreement, we damage 50 fidelity papers (seed 20261101) in graded steps, without any language model. Each paper receives 12 versions: the word order is shuffled within $k$ paragraphs, or $k$ paragraphs are deleted, for $k\in\{1,2,3,5,8,12\}$. The sets are nested, and the paragraphs are body paragraphs of at least 40 words lying wholly inside \Surr's 30,000-character window. The targets are scored with the paper's pipeline; rescoring the unedited originals reproduces their frozen scores exactly. Over the six deletions, the mean within-paper correlations are \CI{0.651}{0.555}{0.742} for \Surr--\Tone, \CI{0.530}{0.391}{0.656} for \Surr--\Ttwo, and \CI{0.470}{0.347}{0.586} for \Tone--\Ttwo{} (Table~\ref{tab:postreview}). The estimand therefore detects agreement when the edits share a graded signal. Deletion also shortens the paper, to which \Surr{} responds through its length features, so this control does not show that any scorer tracks content. Word shuffles separate the targets: shuffling 12 paragraphs lowers \Tone{} by 0.407 SD on average but \Ttwo{} by only 0.049 SD, and \Ttwo's changes barely follow the number of shuffled paragraphs (within-paper correlation with severity \CI{0.056}{-0.133}{0.241}, against \CI{0.465}{0.314}{0.611} for \Tone). Over the shuffles, \Surr--\Tone{} agreement is \CI{0.321}{0.171}{0.464} and \Surr--\Ttwo{} agreement \CI{0.025}{-0.137}{0.185}.

\emph{\Ttwo{} in its own venue.} On the 1,415 papers of its balanced arXiv test split, the split of its model card, our bfloat16 implementation of \Ttwo{} gives an AUC of \CI{0.813}{0.791}{0.834}, accuracy 72.7\%, and a predicted acceptance rate of 46.6\%, against 0.812, 72.8\%, and 47\% reported on its model card \citep{cardarxiv}. We repeat the positive control on 50 of these papers (seed 20261103), with the same damage levels and paragraph rules, measuring changes in SD units of \Ttwo's and \Surr's scores on the 1,415 papers. Over deletions, \Ttwo's changes follow severity at \CI{0.550}{0.427}{0.663} and agree with \Surr{} at \CI{0.546}{0.424}{0.658}, as on ICLR papers (0.526 and 0.530). Over shuffles, they follow severity at \CI{0.187}{0.002}{0.370}, against 0.056 on ICLR papers, and agree with \Surr{} at \CI{0.107}{-0.051}{0.267}; shuffling 12 paragraphs lowers \Ttwo{} by 0.118 SD. Scoring outside its venue therefore does not fully explain \Ttwo's weak response to word order; \Ttwo{} also responds weakly to it on its own test split.

\emph{Equal-size damage.} The graded control rewards a shared response to damage severity. To test equal-size, non-nested changes, each of the same 50 papers receives ten versions, each shuffling the words of one different paragraph (the first ten of the control's seeded order, with the same shuffles), so every version keeps the paper's length. Both targets score them in float16, whose edit changes track 32-bit changes at 0.98 (Section~\ref{app:reliability}); originals reuse the Study 2 float16 scores. Mean within-paper agreement is \CI{-0.068}{-0.144}{0.010} for \Surr--\Tone, \CI{-0.043}{-0.141}{0.057} for \Surr--\Ttwo, and \CI{0.226}{0.128}{0.316} for \Tone--\Ttwo. The mean change is -0.019 SD for \Tone, -0.027 for \Ttwo, and -0.017 for \Surr. \Surr's changes follow the length of the shuffled paragraph (within-paper correlation \CI{-0.280}{-0.370}{-0.190}), the targets' do not (\CI{-0.041}{-0.139}{0.055} and \CI{-0.025}{-0.118}{0.070}). The estimand therefore registers agreement between the two targets on equal-size, non-nested changes, a response that \Surr{} does not share.

\emph{Editable features, edit descriptors, and a learned predictor (post hoc, saved scores).} Of \Surr's 300,008 input columns, 142,086 change in at least one admitted fidelity edit; they carry 57\% of the absolute weight. Restricted to them, \Surr's linear score ranks the 1,667 benchmark papers at 0.668 with \Tone's 32-bit scores and 0.522 with \Ttwo's, against 0.680 and 0.509 for the full linear score; the remaining columns give 0.221 and 0.053. With columns changed by at least ten edits the values are 0.628 and 0.496. The paper-level agreement therefore lies mostly in features the edits can change. For each edit we also measure, on the changed paragraphs, word overlap, the change in sentence count and mean sentence length, connectives, passive constructions, nominalizations, Flesch reading ease, and \Tone's token count. Each correlates with the 32-bit target changes at most 0.06 within papers in absolute value (largest: token-count change with \Ttwo, \CI{0.060}{0.006}{0.114}); four of the 16 intervals exclude zero, all below 0.07. A ridge model on all descriptors, fitted to each target's paper-demeaned 32-bit changes on half of the papers and evaluated on the other half (five splits), reaches \CI{0.004}{-0.025}{0.032} for \Tone{} and \CI{0.025}{-0.011}{0.060} for \Ttwo. Adding sentence embeddings of the old and new paragraphs (bge-small-en-v1.5; difference and new vector) gives \CI{-0.006}{-0.030}{0.019} and \CI{0.008}{-0.025}{0.042}. The same learner recovers \Surr's own changes at only \CI{0.057}{0.023}{0.089}, so it is weak, and these values do not bound what a stronger learner could predict. Fitted in the same way to \Surr's own changes, the ridge model on \Surr's feature changes of Section~\ref{sec:fidelity} recovers them at only \CI{0.234}{0.205}{0.263}, so it does not bound what \Surr's features could predict either.

\emph{Magnitude, between-paper directions, replicate surrogates, and the calibrated scale (post hoc, saved scores).} Within papers, the absolute changes of \Surr{} agree with those of \Tone{} and \Ttwo{} (32-bit) at \CI{0.099}{0.063}{0.134} and \CI{0.101}{0.060}{0.141}, but at \CI{0.027}{-0.013}{0.067} and \CI{0.025}{-0.018}{0.068} after the logarithm of the character change, paragraph position, and edit type are removed with paper fixed effects; the two targets agree on absolute changes at \CI{0.196}{0.159}{0.234}, and at \CI{0.150}{0.107}{0.194} net of these. The targets therefore share which edits move them more than the direction of the move, and \Surr{} shares neither beyond edit size. For directions, we take the 20 leading principal directions of \Surr's centred feature matrix on the 1,667 benchmark papers. The admitted edits place a median of 0.04\% of their squared feature change in that subspace (mean 7.6\%). Splitting each paper's edits at its median share, \Surr--\Tone{} agreement is \CI{-0.069}{-0.123}{-0.013} in the upper half and \CI{0.039}{-0.017}{0.095} in the lower half (\Surr--\Ttwo: \CI{-0.045}{-0.104}{0.013} and \CI{0.009}{-0.044}{0.061}; papers with at least four edits per half), so agreement does not rise along the directions in which papers differ. The ridge model distilled from \Tone's 32-bit paper scores (above) agrees with \Surr{} on edits at \CI{0.208}{0.170}{0.246}, although both are linear in \Surr's features and rank papers alike. Refitting \Surr's recipe (same features, structure weight, and validation choice of the regularization strength) on two disjoint random halves of the 21,141 training papers, for two seeds, gives validation AUCs of 0.67--0.69; the two halves agree with each other on the admitted edits at \CI{0.222}{0.181}{0.262} and \CI{0.200}{0.159}{0.240}, each agrees with \Surr{} at 0.69--0.71, and none agrees with either target (all intervals include zero). On the release's calibrated scale (temperature scaling with the base-rate term), the mean absolute 32-bit edit change is 0.83 points for \Tone{} and 0.55 for \Ttwo, against 1.36 and 1.03 on the raw score, and an added space (the 1,000 twins on 60 papers) moves them by 0.48 and 0.26 calibrated points.

\emph{Agreement expected under a shared component, and the 32-bit AUC gap (post hoc, saved scores).} Suppose a target's standardized score is $\rho\,\Surr + U$, with $U$ independent of \Surr{} and $\rho$ the Pearson correlation of the two scores on the 200 Study~2 originals (32-bit; 0.65 for \Tone{} and 0.41 for \Ttwo). An edit then changes the target by $\rho\,\Delta\Surr + \Delta U$. For each paper we simulate $\Delta U$ as Gaussian noise scaled so that the simulated changes have the target's observed within-paper spread, and compute the mean within-paper Spearman correlation with \Surr's changes over the 198 fidelity papers (1,000 simulations). It is 0.17 [0.13, 0.20] for \Tone{} and 0.23 [0.20, 0.26] for \Ttwo{} (simulation ranges), and 0.17 and 0.27 with the 1,667-paper rank correlations as $\rho$; the observed 32-bit agreement is \CI{-0.008}{-0.047}{0.031} and \CI{-0.021}{-0.059}{0.016}. The model is a reference point, not a fitted account: a target could share \Surr's paper-level direction nonlinearly or only for some edits. On the 1,667-paper benchmark, the paired AUC difference at 32-bit precision is \CI{0.012}{-0.008}{0.034} for \Tone{} minus \Surr{} and \CI{-0.034}{-0.061}{-0.006} for \Ttwo{} minus \Surr{} (2,000 resamples).

\begin{table}[t]
\centering
\small
\setlength{\tabcolsep}{4pt}
\caption{Additional post hoc analyses (descriptive). Agreement is the mean within-paper Spearman correlation \citep{spearman1904} on the 25 rescored papers; pooled and between-paper correlations are Pearson correlations on the 198 fidelity papers. The positive control uses 50 of those papers and six graded damages of each kind; $D$ is the dense surrogate, evaluated on 197 papers. Brackets are nominal 95\% paper-bootstrap confidence intervals.}
\label{tab:postreview}
\begin{tabular}{lll}
\toprule
Quantity & Pair & Estimate [95\% CI] \\
\midrule
Edit agreement, RTX 4090 bfloat16 & \Tone--\Ttwo & \CI{-0.003}{-0.104}{0.101} \\
Edit agreement, RTX 5090 bfloat16 & \Tone--\Ttwo & \CI{0.051}{-0.039}{0.143} \\
Edit agreement, RTX 5090 32-bit & \Tone--\Ttwo & \CI{-0.040}{-0.172}{0.087} \\
Whitespace-edit agreement, RTX 5090 bfloat16 & \Tone--\Ttwo & \CI{0.136}{-0.060}{0.320} \\
Pooled edit changes & \Surr--\Tone & \CI{0.018}{-0.028}{0.067} \\
 & \Surr--\Ttwo & \CI{0.024}{-0.017}{0.066} \\
 & \Tone--\Ttwo & \CI{0.025}{-0.033}{0.084} \\
Per-paper mean changes & \Surr--\Tone & \CI{0.099}{-0.039}{0.253} \\
 & \Surr--\Ttwo & \CI{0.107}{-0.018}{0.231} \\
 & \Tone--\Ttwo & \CI{0.003}{-0.138}{0.142} \\
Positive control, deletions & \Surr--\Tone & \CI{0.651}{0.555}{0.742} \\
 & \Surr--\Ttwo & \CI{0.530}{0.391}{0.656} \\
 & \Tone--\Ttwo & \CI{0.470}{0.347}{0.586} \\
Positive control, word shuffles & \Surr--\Tone & \CI{0.321}{0.171}{0.464} \\
 & \Surr--\Ttwo & \CI{0.025}{-0.137}{0.185} \\
 & \Tone--\Ttwo & \CI{0.150}{0.000}{0.296} \\
Dense surrogate, admitted edits & $D$--\Tone & \CI{0.010}{-0.030}{0.049} \\
 & $D$--\Ttwo & \CI{-0.004}{-0.044}{0.035} \\
\bottomrule
\end{tabular}
\end{table}

\emph{Dense surrogate.} To test a surrogate that reads word order, we fine-tune the sentence encoder bge-small-en-v1.5 \citep{cpack} on the same ICLR training labels. It reads the first 16 chunks of 512 tokens (about 8,000 tokens), pools them with learned attention, and adds the structure counts of \Surr. Training runs for two epochs (batch size 8, learning rate $3\times10^{-5}$, seed 0), and the checkpoint with the best validation AUC is kept. On the 1,667-paper benchmark its AUC is \CI{0.668}{0.642}{0.693}, below \Surr's 0.700. On the 198 fidelity papers its original-paper Spearman correlation is \CI{0.663}{0.561}{0.743} with \Tone{} and \CI{0.370}{0.236}{0.485} with \Ttwo. On the admitted single edits, its mean within-paper agreement is \CI{0.010}{-0.030}{0.049} with \Tone{} and \CI{-0.004}{-0.044}{0.035} with \Ttwo{} (197 papers), and \CI{0.047}{0.009}{0.085} with \Surr. Its edit changes average 0.016 of its SD across the 200 Study 2 originals. In the positive control it agrees with \Tone{} on deletions and on shuffles (0.579 and 0.338), and with \Ttwo{} on deletions only (0.439 and $-0.034$).

\emph{Surrogate fitted to target responses.} For each admitted single edit we take the change in \Surr's own input features (its TF--IDF and scaled structure features, 300,008 columns) and fit a ridge regression to each target's paper-demeaned 32-bit edit change. Papers are split into two halves; the model fitted on one half, with its penalty chosen by an inner split of that half, predicts the edits of the other. Averaged over five random splits, the held-out mean within-paper correlation is \CI{-0.014}{-0.041}{0.012} for \Tone{} and \CI{-0.019}{-0.051}{0.013} for \Ttwo{} (198 papers). A linear model on the surrogate's features, fitted to the targets' own responses, therefore does not recover their edit rankings.

\emph{Response size, formatting, and scale.} Splitting each paper's edits at its median $|\Delta\Surr|$, agreement in the upper half is \CI{0.000}{-0.057}{0.057} with \Tone{} and \CI{-0.044}{-0.102}{0.015} with \Ttwo{} in 32-bit, and \CI{-0.006}{-0.059}{0.045} and \CI{-0.050}{-0.107}{0.005} in bfloat16, so agreement does not rise where \Surr{} moves most. For formatting, we count non-ASCII characters, double spaces, and hyphenated breaks in the changed span of each edit, before and after; 35.8\% of edits change at least one, mostly by removing double spaces. Adding these counts, signed and absolute, to the descriptor regression raises the within-paper $R^2$ of the 32-bit target changes from 0.008 to 0.009 for \Tone{} and from 0.004 to 0.007 for \Ttwo. On the 1,667-paper benchmark, a paired bootstrap (2,000 resamples) gives \Tone{} minus \Surr{} AUC \CI{0.020}{0.001}{0.040} and \Ttwo{} minus \Surr{} \CI{-0.031}{-0.056}{-0.004}. On the release's calibrated scale (the model cards' temperature scaling with the base-rate term), the 32-bit edit changes average 0.058 and 0.041 SD of that scale's originals, against 0.056 and 0.040 on the raw scale, and on the 25 papers of Section~\ref{app:reliability} the whitespace-only null edits move the targets by 45\% and 47\% of the real-edit change, against 45\% and 50\% raw.

\FloatBarrier

\appendixsection{Generative reviewer evaluation}
\label{app:deepreviewer}
\label{sec:deepreviewer}

We additionally evaluate 24 Study 3 papers with DeepReviewer-14B \citep{deepreviewer}, a Phi-4-based generative model \citep{phi4} used here in quantized form. Each original, adaptive endpoint at 512, random endpoint at 512, and polished version receives two Fast Mode reviews. The mean original rating is 5.97 on its 1--10 scale; the between-paper SD is 0.935, and the mean within-text SD across paired reviews is 0.425 rating points (91 texts with two ratings).

Adaptive minus random change is \CI{-0.02}{-0.32}{0.27} rating points on 19 paired papers. The polishing change is \CI{0.12}{-0.09}{0.33} on 24 papers. This small descriptive check does not show a search advantage, but its intervals allow moderate effects of approximately 0.3 between-paper SD in either direction. It therefore provides only limited cross-family evidence. Six of 192 reviews reach the output cap without a rating; seven edited versions that fall back to the original are excluded from change estimates. 

The DeepReviewer check selects the first 24 papers in the seeded Study 3 sample. It uses the original, adaptive endpoint at 512 with search seed zero, random endpoint at 512 with search seed zero, and filtered one-pass polish. Each version is reviewed with generation seeds one and two, temperature 0.4, top-$p$ 0.95, a 14,000-token input limit, and a 6,000-token output cap. The model is the quantized DeepReviewer-14B listed in the version table, run with llama.cpp \citep{llamacpp}. 

\noindent\begin{minipage}{\linewidth}
The Fast Mode system message is:
\begin{quote}\small
You are an expert academic reviewer tasked with providing a thorough and balanced evaluation of research papers. Your thinking mode is Fast Mode. In this mode, you should quickly provide the review results.
\end{quote}
\end{minipage}\par

Ratings are extracted from the structured review output and averaged within paper and version over available rated reviews. Of 192 attempted reviews, 186 contain a rating; six hit the output cap without one. Seven of the 72 edited versions have no usable checked endpoint and fall back to the original, producing 14 fallback reviews. These versions are excluded from the corresponding change estimates, not counted as zero-gain versions. This differs deliberately from the primary transfer estimand, where invalid endpoints count as zero gain.

\begin{table}[htbp]
\centering
\caption{DeepReviewer-14B descriptive changes on its 1--10 rating scale. These are rating points, not the standardized units used for the acceptance predictors. Confidence intervals (CIs) bootstrap papers; available seeds are averaged within paper.}
\label{tab:deepreviewer}
\begin{tabular}{lrc}
\toprule
Contrast & Paired papers & Mean change [95\% CI] \\
\midrule
Adaptive minus original & 19 & \CI{-0.008}{-0.312}{0.312} \\
Random minus original & 21 & \CI{0.017}{-0.218}{0.247} \\
Polish minus original & 24 & \CI{0.116}{-0.090}{0.329} \\
Adaptive minus random & 19 & \CI{-0.021}{-0.317}{0.268} \\
\bottomrule
\end{tabular}
\end{table}

There are 65 non-fallback edited versions; nine exceed the input limit. Of their changed paragraphs, 111 of 112 adaptive changes, 120 of 122 random changes, and 672 of 687 polish changes fall inside the retained input. Coverage is therefore approximately 98--99\%, rather than complete. These coverage counts refer to candidate versions before rating missingness, explaining why 20 adaptive versions are available for input inspection while only 19 enter a rated comparison. There are 24 truncated review calls across all versions and seeds, including originals. The narrow main-study null should not be attributed to this smaller, noisier evaluation.

\emph{Edit-level readout.} After the main analyses, we also score inputs with DeepReviewer-14B without generating reviews. Generating one review for each of the 3,523 Study~2 originals and single-edit versions alone would take about 55 hours at the observed mean of 56 seconds per review. The prompt keeps the Fast Mode system message and the model's own opening of a Fast Mode answer, then jumps to the answer's final fields. The primary readout, fixed on a 30-paper pilot before any edit was scored, is the expected leading digit of the rating after ``\texttt{\#\# Rating:}''; the secondary readout is the log-probability difference between Accept and Reject after ``\texttt{\#\# Decision:}''. Inputs are cut to 14,000 tokens as above. We score a seeded, label-balanced subset of 300 benchmark papers, the 200 Study~2 originals and 3,323 single edits, the 1,000 added-space versions of 60 papers, and the 500 single-paragraph scrambles of 50 papers. Changes are standardized by the readout's SD over the 200 originals (0.32 rating points). A first run, in which the server reused computation cached from the previously scored input, made scores depend on scoring order; we discarded it. In the reported run, rescoring the 200 originals in reverse order reproduces every score exactly.

The rating readout separates accepted from rejected papers with an AUC of \CI{0.679}{0.620}{0.738} and ranks the 300 papers with \Surr, \Tone, and \Ttwo{} at Spearman 0.518, 0.531, and 0.330 (32-bit target scores). On the 198 fidelity papers, its edit changes agree with \Surr's at \CI{0.003}{-0.034}{0.041}, with \Tone's at \CI{-0.012}{-0.050}{0.026}, and with \Ttwo's at \CI{0.022}{-0.014}{0.060}. A score that contained \Surr's at the observed correlation on the originals (Pearson 0.557) would, given the observed spread of both scorers' changes, agree with \Surr{} at about 0.099 (simulation interval 0.061 to 0.137; Supplementary Section~\ref{app:postreview}); the upper limit of 0.041 lies below this. The 25 papers longer than the input limit contain 14 edits outside it; each changes the score by exactly zero, and excluding them moves each estimate by less than 0.003. DeepReviewer's edit responses are less stable than the targets': an added space moves the readout by 0.062 SD against 0.079 SD for the same edits, and the two changes correlate within papers at \CI{0.370}{0.314}{0.427}. Under independent errors, this limits agreement to about 0.60 with \Surr{} and 0.53 with each target, far above the observed values. The positive control does not hold, however. The single-paragraph scrambles move the readout by 0.113 SD, yet its changes agree with neither target's (\CI{-0.020}{-0.104}{0.066} and \CI{-0.025}{-0.110}{0.055}), whereas the targets agree with each other at \CI{0.226}{0.128}{0.316}. DeepReviewer therefore shares the targets' response to none of the paragraph-level changes tested here, and this check cannot separate an absence of shared edit preferences from a readout that ranks such changes differently. The secondary decision readout separates papers weakly (AUC \CI{0.562}{0.494}{0.626}) and agrees with no scorer on edits either (means from $-0.024$ to 0.003).

\appendixsection{Protocol chronology and reproducibility}
\label{app:protocol}
The executed study uses seed 20261029 for study sampling and audit sampling, with pilot seed 20261028, search seeds zero and one, and generative-reviewer seeds one and two. Study sampling excludes all 30 pilot papers. The operational sequence is edit generation, invariant and semantic checking, surrogate-only search, complete endpoint checks at 64 and 512, sampled second-provider checking, the corruption challenge, and freezing. Target scoring follows the freeze. The human audit and disagreement adjudication follow scoring; neither changes the four primary tests. The exploratory best-edit analysis is added after examining the main outcomes.

\begin{table}[htbp]
\centering
\caption{Paper cohorts and nested analysis sets. The single-edit fidelity set is a subset of the admitted-edit pool; the visible-span analysis retains the same fidelity papers with fewer edits. One selected Study 2 paper has no admitted single edit. Rows overlap and must not be added.}
\label{tab:cohorts}
\begin{tabularx}{\linewidth}{@{}Xr>{\raggedright\arraybackslash}X@{}}
\toprule
Set & Papers & Inputs or inclusion rule \\
\midrule
Study 2 selected & 200 & All retained as originals \\
Admitted single-edit pool & 199 & 3,323 edited versions \\
Edit fidelity & 198 & 3,321 edits; at least five per paper and varying scores \\
Visible-span fidelity & 198 & 3,072 edits; complete changed paragraphs visible to all scorers \\
Paired deletion diagnostic & 93 & One admitted removal and sham per paper \\
Study 3 search & 128 & Same papers under both policies and seeds \\
Standardization cohort & 328 & The 200 Study 2 and 128 Study 3 originals \\
\bottomrule
\end{tabularx}
\end{table}

There are 5,321 inputs: 328 originals, 3,323 single-edit versions, 102 evidence removals, 184 deletion controls, 128 polished versions, 683 checked composite endpoints, and 573 exploratory endpoints from the two smallest search budgets. The saved input list records every text and its order. Before target scoring, the program checks each text and the complete ordered list against the saved versions. These checks establish that the evaluation inputs remain unchanged; they do not establish public preregistration. Both targets produce finite scores for all 5,321 inputs.

\begin{table}[H]
\centering
\small
\caption{Versions of the evaluated models, data, and software. Exact release identifiers and checks for file changes are recorded with the planned code release (Data and code availability).}
\label{tab:versions}
\begin{tabularx}{\linewidth}{@{}p{0.22\linewidth}X@{}}
\toprule
Component & Identifier and version \\
\midrule
\Tone & \texttt{skonan/PaperScorer-3B-Text-OpenReview-ICLR}, revision \texttt{cc25000d6751} \\
\Ttwo & \texttt{skonan/PaperScorer-3B-Text-arXiv}, revision \texttt{5a25bebf9dcc} \\
Source text & \texttt{skonan/PaperScorer-Text}, revision \texttt{c39ba04eb7d1} \\
Target inference & PyTorch \citep{pytorch} 2.11.0 (CUDA 12.8, NVIDIA's GPU computing platform), Transformers \citep{transformers} 5.21.0, Tokenizers 0.23.3; bfloat16 forward pass, 32-bit Accept/Reject logits \\
\Surr & scikit-learn \citep{sklearn} 1.9.1 \\
DeepReviewer-14B & \texttt{WestlakeNLP/DeepReviewer-14B}; converted to the GGUF model-file format with Q8\_0 (8-bit) quantization; served by llama.cpp release b11972, with a converter patch mapping the model's tokenizer to llama.cpp's cl100k-style (\texttt{dbrx}) pre-tokenizer \\
Writer and checkers & GPT-6-Luna and GPT-6.1-Sol through the Codex command-line interface (CLI) 0.165.0, called on 28--29 October 2026; Claude Opus~5.5 (\texttt{claude-opus-5-5}) through Claude Code \\
\bottomrule
\end{tabularx}
\end{table}

The main study completes 456 writer tasks, 456 primary-checker tasks, 128 endpoint-checking tasks, 20 corruption-checking tasks, and one corruption-generation task: 1,061 tasks through the OpenAI-based tooling. Four tasks require one retry each, giving 1,065 logged attempts. The second-provider sample completes 59 tasks without retries. Recorded usage for the successful OpenAI-based attempts, excluding the second-provider sample, is approximately 37.68 million input tokens, including 15.46 million cached tokens, and 9.17 million output tokens. Their summed durations are approximately 28.77 hours, with parallel execution reducing wall time. Failed-attempt usage and duration are not retained and are excluded from these totals. These are study-generation and validation costs, not part of the 7-millisecond surrogate inference claim. They also exclude the pilot, artificial intelligence (AI) assistance with this article, and the generative-reviewer check. Each acceptance predictor scores the frozen set in approximately 2.1 hours on the RTX 4090. AI assistants supported implementation of the analysis and plotting scripts.

The initial design included more papers, whole-paper rewrite conditions, a human-validity gate before scoring, and a human-reviewer anchor. The executed design reduces the samples, drops whole-paper rewrite conditions after pilot failures, measures human validity after scoring, and omits the human anchor because suitable version-matched histories were not assembled. These departures narrow the claims and are reported rather than treated as completed protocol elements. None of the four primary tests is selected because its effect is positive.

For independent reconstruction, the essential records are source paper identifiers and split membership; the fixed family-filter rules and sampling seeds; the writer and checker model configurations; edit positions and admission decisions; search constraints and policy seeds; records of the exact input versions; evaluator prompts, truncation, and precision; scores; and paper-level aggregation rules. The planned release preserves these non-text records and analysis code while excluding source and edited manuscript text. Reobtaining model weights and source text remains subject to their original access and licensing terms.

\end{document}